\documentclass[12pt]{article}
\usepackage[utf8]{inputenc}
\usepackage[T1]{fontenc}

\usepackage[compress]{cite}
\usepackage{IEEEtrantools,stackengine}
\stackMath

\usepackage{authblk}
\usepackage{fullpage}
\usepackage{amssymb,amsmath}
\usepackage{adjustbox}
\usepackage{graphicx}
\usepackage{longtable}
\usepackage{pdfpages}

\usepackage{siunitx}
\usepackage{csquotes}

\usepackage{xspace}

\usepackage{booktabs}
\usepackage{longtable}
\usepackage{multirow}

\usepackage[dvipsnames, table]{xcolor}
\definecolor{dark green}{rgb}{0.0,0.5,0.0}

\definecolor{darkgreen}{rgb}{0.0, 0.5, 0.0}

\definecolor{lightyellow}{HTML}{FFE699}
\definecolor{red_revision}{HTML}{FF0000}

\usepackage{setspace}
\usepackage[unicode=true]{hyperref}
\hypersetup{breaklinks=true,
            pdfborder={0 0 0},
            allcolors=black}
\usepackage[%
  sort&compress
]{cleveref}
  \Crefname{appendix}{Supplement}{Supplements}
  \Crefname{figure}{Fig.}{Fig.}

\usepackage{times}
\usepackage{enumerate}
\usepackage{enumitem}
\usepackage{float}

\usepackage{subfig}
\usepackage{graphicx}
\usepackage{longtable}

\usepackage{rotating}
\usepackage{tabularx}
\usepackage{dcolumn}
\usepackage{pdflscape}
\usepackage{rotating}

\usepackage{array}

\usepackage{titlesec}

\titleformat{\subsubsection}
  {\normalfont\itshape}{\thesubsubsection}{1em}{}

\titleformat{\subsection}
  {\normalfont\bfseries}{\thesubsection}{1em}{}

\makeatletter
\renewcommand{\fps@figure}{H}         
\renewcommand{\fps@table}{H}         
\makeatother

\allowdisplaybreaks

\usepackage{eurosym}
\usepackage{comment}

\usepackage{ntheorem}
\theoremseparator{:}

\usepackage{makecell}

\usepackage{listings}
\usepackage[most]{tcolorbox}
\tcbuselibrary{listings,breakable,skins}
\usepackage{caption}
\usepackage{inconsolata}   
\usepackage{sourcecodepro}
\usepackage[scaled]{beramono}

\lstdefinestyle{promptstyle}{%
  basicstyle=\ttfamily\scriptsize,
  breaklines=true,
  breakindent=1.5em,
  breakatwhitespace=true,
  columns=fullflexible,
  keepspaces=true,
  showstringspaces=false,
  upquote=true,
  aboveskip=0pt,
  belowskip=0pt,
  xleftmargin=0pt,
  framexleftmargin=0pt,
}
 
\newtcblisting{promptbox}[1]{%
  listing only,
  listing options={style=promptstyle},
  breakable,
  enhanced jigsaw,
  colback=black!3,
  colframe=black!55,
  boxrule=0.4pt,
  arc=1.5pt,
  left=4pt, right=4pt, top=3pt, bottom=3pt,
  title=#1,
  fonttitle=\bfseries\footnotesize\rmfamily,
  colbacktitle=black!14,
  coltitle=black,
  width=\linewidth,
  before skip=6pt,
  after skip=11pt,
}

\usepackage{outlines}
\usepackage{pifont}                 
\usepackage[section]{placeins}      

\begin{document}


\title{\centering\LARGE\singlespacing Interactions with large language models increase support for refugees}

\renewcommand\Affilfont{\fontsize{9}{10.8}\selectfont}

\author[1,$\dagger$]{Amirsiavosh Bashardoust}
\author[2,$\dagger$]{Ksenia Bykanova}
\author[2,3]{Sebastian Maier}
\author[2,3]{Abdurahman Maarouf}
\author[2,3]{Dominique Geissler}
\author[1]{Yash Raj Shrestha}
\author[2,3]{Stefan Feuerriegel\thanks{Corresponding author: feuerriegel@lmu.de}}

\affil[$\dagger$]{These authors contributed equally to this work.}

\affil[1]{University of Lausanne (UNIL), HEC Lausanne, Lausanne, Switzerland}
\affil[2]{LMU Munich, Munich, Germany}
\affil[3]{Munich Center for Machine Learning, Munich, Germany}

\date{}

\maketitle

\newpage

\begin{abstract}\normalfont
\noindent

\noindent
Public support for refugees remains limited across many countries, and, in particular, scalable strategies to increase support are still scarce. Here, we test whether brief conversations with large language models (LLMs) can positively shift attitudes toward refugees and related policy support. For this, we introduce a set of dialogue-based LLM interventions designed to promote reflection, empathy, and perspective-taking; these include: (1) an interview-style interaction to encourage self-reflection, (2) an empathy-focused dialogue, and (3) an immersive roleplay. For comparison, we also test (4) a debunking condition that provides factual corrections. We evaluate these interventions across three large, preregistered experiments ($N = 1358$ in total). In Study 1 ($N = 554$, U.K.), we find significant within-subject effects for both the empathy-focused and the roleplay intervention. For example, the roleplay intervention increased support for refugee support by $9.7$ percentage points (p.p.) and policy endorsement by $5.7$ p.p. (both $p < 0.001$). In Study 2 ($N = 250$, Kenya), we replicate these effects in a non-WEIRD country, again observing that empathy and roleplay are effective. In Study~3 ($N = 554$, U.K.), we examine long-term effects using a between-subject design. Together, these findings demonstrate that brief, LLM-based dialogues can serve as scalable interventions to increase support for refugees. Moreover, the results provide insights into the mechanisms underlying these effects and provide recommendations for designing effective interventions that promote more supportive attitudes toward refugees.
\end{abstract}

\flushbottom
\maketitle
\thispagestyle{empty}


\sloppy
\raggedbottom


\newpage

\section*{Main}


An estimated 123.2 million people were forcibly displaced worldwide, equivalent to roughly one in 67 people \cite{UnitedNationsHighCommissionerforRefugees.2025}. In 2024, OECD countries registered a record 3 million new asylum applications, alongside increases in humanitarian migration and refugee resettlement \cite{OECD.2025}. At the same time, negative attitudes toward refugees are widespread, with immediate political and social consequences. Large segments of the public oppose accepting refugees, which has contributed to the electoral success of anti-refugee campaigns and increased support for far-right parties \cite{Dinas.2019, Hainmueller.2014}. Negative attitudes are also associated with local hostility, including protests and incidents of violence against refugee accommodations \cite{Dinas.2019}.

However, attitudes toward refugees are difficult to change because they are closely tied to religious, ethnic, and ideological identities \cite{Hainmueller.2014, Hopkins.2019}. One reason is that attitudes toward refugees are rooted in sociotropic concerns and perceptions of symbolic threat, meaning that individuals often evaluate refugees in terms of perceived national consequences, such as economic burden or cultural cohesion, rather than personal experience \cite{Hainmueller.2014}. Because such concerns are embedded in broader beliefs and worldviews, promoting more supportive attitudes toward refugees through behavioral interventions requires a careful understanding of the underlying psychological mechanisms.


Research in social and political psychology has identified several mechanisms that can potentially promote more positive attitudes toward refugees. (1)~One mechanism is \textit{self-reflection}, which draws on cognitive dissonance and self-persuasion. The idea is that, by prompting individuals to articulate their own reasons and values, one can reduce defensiveness and encourage more internally consistent attitudes \cite{Aronson.1999, Argyle.2025}. (2)~A second mechanism is \textit{empathy}, based on intergroup emotion theory, whereby adopting the viewpoint of an outgroup member can reduce psychological distance and increase prosocial motivation \cite{Batson.1997, Todd.2014}. (3)~A third mechanism is \emph{perspective-taking}, which encourages identity reframing by placing individuals in first-person scenarios, thereby increasing engagement and reducing defensive processing \cite{Green.2000, Kalla.2023, Adida.2018}. (4)~A fourth mechanism is \textit{debunking}, which aims to correct misconceptions by providing evidence-based counterarguments and factual information to challenge existing beliefs \cite{Costello.2024, Rabb.2025}. However, evidence on the effectiveness of factual corrections in changing attitudes is mixed, particularly when attitudes are closely tied to broader worldviews and social identities \cite{Ecker.2024, Walter.2018, Hopkins.2019}. This suggests that factual arguments alone may be insufficient to change attitudes that are moralized, emotionally charged, and linked to group identity \cite{Hopkins.2019, Kubin.2021}. As a result, the above mechanisms---namely, self-reflection, empathy, perspective-taking, and debunking---may vary in their effectiveness to promote more supportive attitudes toward refugees.


Recent advances suggest that LLM-based conversational dialogues can influence attitudes across a range of domains \cite{Maier.2026, Matz.2024}. Short conversations with LLMs have been shown to reduce belief in conspiracy theories \cite{Costello.2024} and to counter both broad conspiracy narratives and identity-linked antisemitic claims \cite{Rabb.2025}. Other work shows that interactive AI dialogues can decrease inaccurate election denial beliefs \cite{Hopkins.2026}. LLM-based interventions have also proven effective in public health communication \cite{Karinshak.2023}, where such interventions can increase intentions for vaccinations \cite{Xu.2025} and improve attitudes toward COVID-19 immunization by addressing individual concerns and misperceptions \cite{Altay.2023}. Further, LLM-based interventions can also be used to shape political preferences \cite{Argyle.2023,Lin.2025,Bai.2025,Breum.2023,Czarnek.2025b,Hoelbling.2025}. However, it is unclear whether the findings from prior research generalize to attitudes toward refugees. Some attitudes, especially those that are identity-relevant, morally charged, or socially embedded, may respond differently to conversational persuasion than more weakly held or informational beliefs \cite{Baer.2024}. This motivates our research, in which we develop and test LLM-based interventions grounded in the above psychological mechanisms to promote more supportive attitudes towards refugees.


Using LLM-based interventions for this task offers several advantages. (1)~LLM-based interventions are scalable. Once deployed, LLM systems can reach large and diverse populations at relatively low marginal cost. In particular, unlike traditional media campaigns \cite{Lin.2025} or resource-intensive interventions such as door-to-door canvassing \cite{Kalla.2023}, LLM-based approaches can be implemented at scale across different contexts, such as the Global South, which is otherwise harder to reach through conventional interventions. Such interventions could be deployed by non-governmental organizations (NGOs), public agencies, and other institutions seeking to promote more supportive attitudes toward refugees. (2)~LLMs offer new opportunities for personalization by tailoring interventions to individual users \cite{Matz.2024}. In the context of attitudes towards refugees, this enables fine-grained adjustments of message content and style to individual concerns, values, and conversational cues. (3)~LLMs also offer a standardized setting to test different persuasion strategies in a controlled and replicable setting. This makes it possible to compare different persuasion mechanisms within a unified framework and to generate insights into how mechanisms of attitude change operate in practice.

\begin{figure}[H]
    \centering
    \vspace{-1cm}
    \includegraphics[width=.95\textwidth, trim=0 0 0 0, clip]{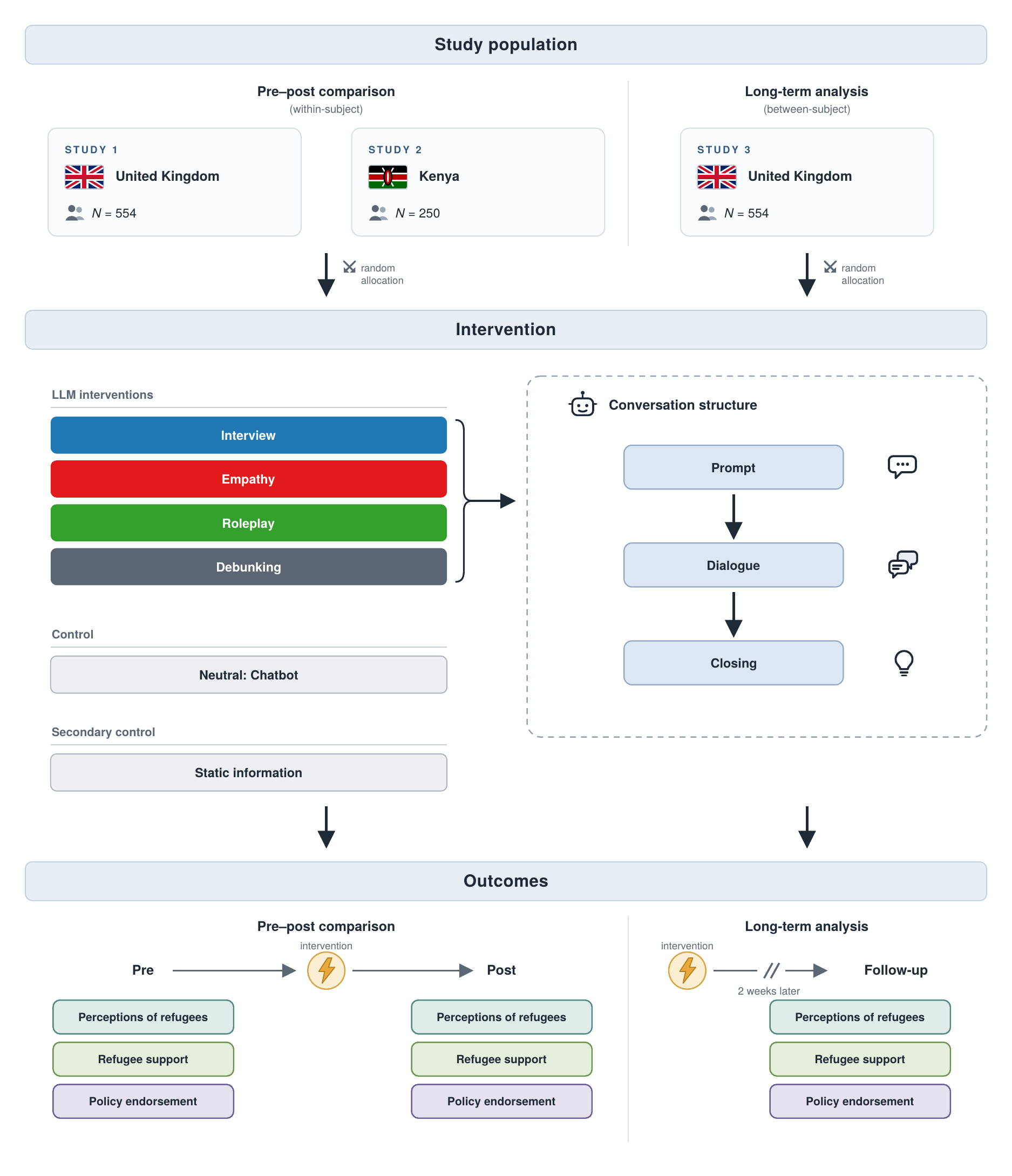}
\vspace{-1cm}

\caption{\textbf{Research design.} Participants were randomized to different LLM-based interventions: \textsc{Interview} to promote self-reflection, \textsc{Empathy} to promote empathy, \textsc{Roleplay} to promote perspective-taking, \textsc{Debunking} to provide factual corrections to existing beliefs. We also compare control condition using a \textsc{Neutral} chatbot with a dialogue about daily life activities and a secondary control \textsc{Static Information} that presents facts in form of a flyer with evidence-based arguments but no interactive dialogue. Study 1 (U.K.; $N = 554$) and Study 2 (Kenya; $N = 250$) used a pre--post design. Study 3 (U.K.; $N = 554$) used a longitudinal between-subject design with a two-week follow-up. Changes are measured along three dimensions: general perceptions of refugees, refugee support, and policy endorsement. }
\label{fig:study_overview}
\end{figure}

In this paper, we introduce and test brief, dialogue-based LLM interventions designed to promote positive attitudes toward refugees and increase support for refugee-related policies (see \Cref{fig:study_overview}). Our interventions compare different psychological mechanisms: (1)~an interview-style interaction to encourage self-reflection (\textsc{Interview}), (2)~an empathy-focused dialogue (\textsc{Empathy}), and (3)~an immersive roleplay to encourage perspective-taking (\textsc{Roleplay}). For comparison, we also report a debunking intervention that provides factual corrections (\textsc{Debunking}),  test one control condition with a neutral LLM chatbot (\textsc{Neutral}), and one reference condition with static information in the form of a flyer with text-based arguments (\textsc{Static information}). We evaluate the interventions across three large-scale preregistered experiments ($N = 1358$ in total). Study~1 ($N = 554$, U.K.) uses a within-subject experiment design to assess pre–post changes in attitudes. Study~2 ($N = 250$, Kenya) replicates the experiment in a non-WEIRD context (i.e., outside Western, Educated, Industrialized, Rich, and Democratic populations). Thereby, we follow calls in prior research to extend findings beyond WEIRD populations \cite{Henrich.2010, Ghai.2025, Rad.2018} and provide a direct test of how the findings generalize in a setting where refugee presence and related societal pressures are particularly salient. Of note, Kenya is among the countries hosting the largest number of refugees worldwide, with more than $600{,}000$ displaced individuals residing in camps such as Dadaab and Kakuma \cite{UNHCR.2026}. Study~3 ($N = 554$, U.K.) examines long-term effects in a between-subject design with a two-week follow-up. In all studies, we measured changes across three outcomes, namely, general perceptions of refugees, refugee support, and policy endorsement.


Across the three experiments, brief LLM-based dialogues led to increases in support for refugees and related policies. In particular, the \textsc{Roleplay} intervention, based on perspective-taking through an immersive approach, showed the largest effects. For example, in Study~1, the \textsc{Roleplay} intervention increased refugee support by $9.7$ percentage points (p.p.) and policy endorsement by $5.7$ p.p. (both $p < 0.001$). These patterns replicated in Study~2 conducted in Kenya, thereby demonstrating that the effects generalize beyond Western contexts. Study~3 found weaker but directionally consistent effects ($p = 0.070$), suggesting that sustained attitude shifts may require repeated engagement. Even though the support for refugees increased, actual perceptions of refugees remained stable, suggesting that evaluations of the latter type are associated with relatively stable attitudes. Together, these findings indicate that short, dialogue-based interactions with LLMs can serve as scalable interventions to promote more supportive attitudes toward refugees, while also offering insights into the psychological mechanisms that are effective for driving attitude change in a refugee context.

\clearpage

\clearpage

\section*{Results}
\label{sec:results}

\subsection*{LLM interventions to promote support for refugees}

We develop and test three LLM-based interventions designed around different psychological mechanisms (\Cref{fig:study_overview}; see Methods for details): (1) self-reflection (\textsc{Interview}), (2) empathy-focused dialogue (\textsc{Empathy}), (3) perspective-taking using immersive roleplay (\textsc{Roleplay}). The \textsc{Interview} intervention prompts participants to articulate and examine their own views through an interview-style dialogue; the \textsc{Empathy} intervention encourages participants to imagine the lived experiences of refugees; the \textsc{Roleplay} intervention places participants in a first-person narrative scenario. For comparison, we also report a fourth LLM-based intervention based on factual debunking (\textsc{Debunking}). The \textsc{Debunking} intervention provides corrective information that addresses misconceptions about refugees. In addition, we include a \textsc{Neutral} chatbot as control condition that consists of a short, LLM-based dialogue about everyday topics with no refugee-related content, while the \textsc{Static Information} acts as a secondary control  that presents evidence-based arguments about refugees as a static flyer, but without interactive dialogue. \textsc{Static Information} thus helps to isolate the effect of the interactive format; we report all results in comparison to the reference condition in Supplement~\ref{suppsec:reanalyis_static_information}. In the following, our main comparison is with respect to the \textsc{Neutral} chatbot.

We assess three outcomes: (1)~general perceptions of refugees (e.g., perceived threat and general evaluations), (2)~refugee support (e.g., willingness to provide support and perceived responsibility), and (3)~policy endorsement (e.g., support for refugee-related policies). The items for measuring general perceptions of refugees are adapted from \cite{Hangartner.2019}, and are recorded on 5-point Likert scales consistent with the original paper. Refugee support and policy endorsement are measured as multi-item indices on 0--100 sliders, following the response format used for policy support and behavioral intentions in recent intervention studies \cite{Vlasceanu.2024, Czarnek.2025b}, where finer resolution is preferred for graded, quantitative judgments. Effect sizes were estimated using linear mixed-effects models; we report the corresponding standard error (SE) and 95\% confidence interval (CI).

\subsection*{Study 1 (United Kingdom)}


Study~1 was conducted as a preregistered online experiment with $N = 554$ participants from the U.K. to evaluate the effects of our four LLM-powered interventions. The study followed a within-subjects design to compare pre--post outcomes. Overall, 39\% of participants were aged 18--34, 44\% aged 35--54, and 17\% aged 55 or older. The gender distribution was fairly balanced (55\% female, 45\% male, and < 1\% identifying as another gender identity). Participants reported moderately left-leaning political orientations ($M = 44.54$, SD $= 25.63$) and low-to-moderate prior contact with refugees ($M = 13.36$, SD = $18.23$).

\paragraph*{Perceptions of refugees.}

\begin{figure}[H]
    \centering
    \vspace{-1cm}
    \includegraphics[height=0.8\textheight]{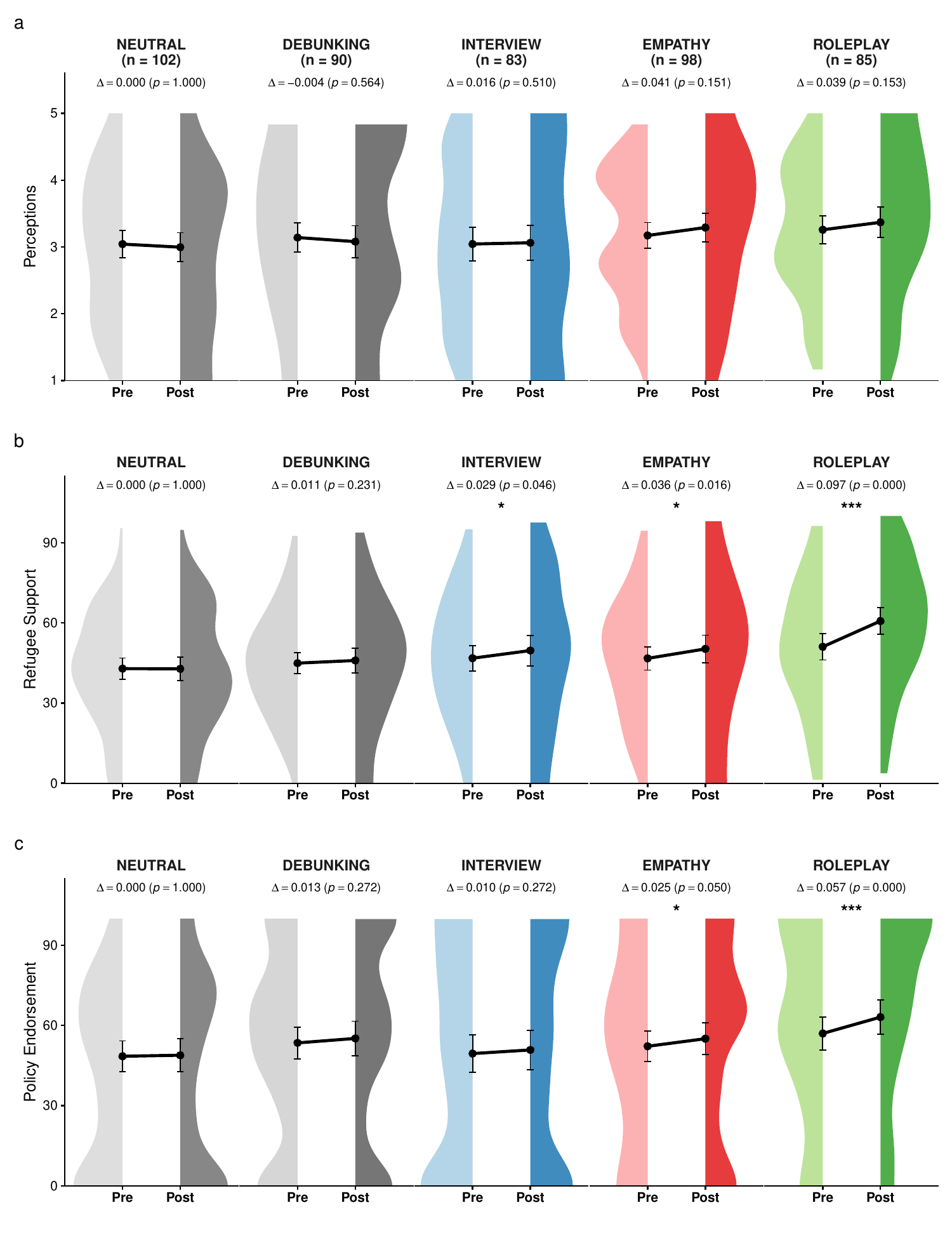}
\captionof{figure}{\textbf{Study 1 ($N = 554$) in the U.K. to estimate the effect of LLM interactions on refugee-related outcomes.} Each panel compares the pre- and post-intervention measurements for the different experimental conditions (i.e., \textsc{Neutral}, \textsc{Debunking}, \textsc{Interview}, \textsc{Empathy}, \textsc{Roleplay}). The estimated changes were analyzed using linear mixed-effects models with condition, time, and their interactions as fixed effects. The \textsc{Neutral} condition is the reference level. Violin plots show the distribution of raw measurements. The black dots show the mean, and whiskers the 95\% confidence intervals. The estimated effects ($\Delta$) represent the additional pre–post change in each intervention relative to the \textsc{Neutral} condition, where $\Delta = 0.01$ corresponds to an increase of 1 percentage point (p.p.). The $p$-values refer to Holm-corrected one-sided tests of the condition $\times$ time interaction. Significance levels: $^{\dagger} p < 0.1$, $^{*} p < 0.05$, $^{**} p < 0.01$, $^{***} p < 0.001$.}
\label{fig:Confirmatory_Stacked_Figure}
\end{figure}

The LLM interventions did not affect the participants' perceptions of refugees (see \Cref{fig:Confirmatory_Stacked_Figure}a). After Holm correction, no contrast differed significantly from the \textsc{Neutral} control (all $p > 0.15$). The estimated changes were: \textsc{Debunking}: $-0.4$ p.p. ($b = -0.004$, SE $= 0.024$, $t = -0.162$, $p = 0.565$); \textsc{Interview}: $1.6$ p.p. ($b = 0.016$, SE $= 0.024$, $t = 0.660$, $p = 0.510$); \textsc{Empathy}: $4.1$ p.p. ($b = 0.041$, SE $= 0.023$, $t = 1.781$, $p = 0.151$); and \textsc{Roleplay}: $3.9$ p.p. ($b = 0.039$, SE $= 0.024$, $t = 1.640$, $p = 0.153$).

\paragraph*{Refugee support.}

Several interventions had large effects on refugee support (see \Cref{fig:Confirmatory_Stacked_Figure}b). The \textsc{Debunking} intervention led to no statistically significant change ($1.1$ p.p.; $b = 0.011$, SE $= 0.014$, $t = 0.735$, $p = 0.231$). The \textsc{Interview} intervention led to a small but statistically significant increase after Holm correction ($2.9$ p.p.; $b = 0.029$, SE $= 0.015$, $t = 2.000$, $p = 0.046$). The \textsc{Empathy} intervention produced a significant increase ($3.6$ p.p.; $b = 0.036$, SE $= 0.014$, $t = 2.573$, $p = 0.016$), and the \textsc{Roleplay} intervention produced the largest increase ($9.7$ p.p.; $b = 0.097$, SE $= 0.015$, $t = 6.651$, $p < 0.001$). As an exploratory analysis, we performed pairwise comparisons between interventions (see Supplement~\ref{suppsec:s1_pairwise}). \textsc{Roleplay} improved refugee support more than \textsc{Debunking} ($8.7$ p.p., 95\% CI: $[5.7, 11.6]$, $p< 0.001$), \textsc{Interview} (difference in change: $6.8$ p.p., 95\% CI: $[3.8, 9.8]$, $p < 0.001$), and \textsc{Empathy} ($6.1$ p.p., 95\% CI: $[3.2, 9.0]$, $p < 0.001$), whereas the other three interventions did not differ from one another (all $p \geq 0.235$), implying that perspective-taking performed best overall.

\paragraph*{Policy endorsement.}

A similar pattern is observed for policy endorsement (see \Cref{fig:Confirmatory_Stacked_Figure}c). The \textsc{Debunking} intervention ($1.3$ p.p.; $b = 0.013$, SE $= 0.012$, $t = 1.100$, $p = 0.272$) and the \textsc{Interview} intervention ($1.0$ p.p.; $b = 0.010$, SE $= 0.012$, $t = 0.812$, $p = 0.272$) did not alter endorsement at a statistically significant level. The \textsc{Empathy} intervention produced a small but significant increase after Holm correction ($2.5$ p.p.; $b = 0.025$, SE $= 0.011$, $t = 2.138$, $p = 0.050$). In contrast, the \textsc{Roleplay} intervention produced a significant increase ($5.7$ p.p.; $b = 0.057$, SE $= 0.012$, $t = 4.824$, $p < 0.001$). As an exploratory analysis, we again performed pairwise comparisons between interventions (see Supplement~\ref{suppsec:s1_pairwise}): \textsc{Roleplay} improved policy endorsement more than \textsc{Debunking} (difference in change: $4.5$ p.p., 95\% CI: $[2.0, 6.9]$, $p = 0.002$); \textsc{Interview} ($4.8$ p.p., 95\% CI: $[2.3, 7.2]$, $p = 0.001$), and \textsc{Empathy} ($3.3$ p.p., 95\% CI: $[0.9, 5.7]$, $p = 0.026$); no other pair of interventions differed (all $p \geq 0.666$).

\paragraph*{Robustness checks.} To assess whether the results depend on the choice of reference category, we re-estimated all models using the \textsc{Static Information} instead of the \textsc{Neutral} intervention as the baseline, while we kept the model specification and outcome variables unchanged. Across all three outcomes, the main findings remained consistent (see Supplement~\ref{suppsec:reanalyis_static_information}). (1)~For the general perceptions of refugees, the estimates were not statistically significant. (2)~For refugee support, the results also remained robust. \textsc{Roleplay} again showed clear improvement relative to the static information condition ($b = 0.075$, $p < 0.001$). \textsc{Empathy} and \textsc{Interview} showed smaller, non-significant effects ($p = 0.604$ and $p = 0.898$, respectively). (3)~For policy endorsement, \textsc{Roleplay} again showed a positive change but slightly above the preregistered significance level  ($b = 0.028$, $p = 0.055$ after Holm correction). Taken together, the \textsc{Roleplay} effect on refugee support is robust to the choice of reference condition.

\subsection*{Study 2 (Kenya)}

Study~2 is a preregistered, direct conceptual replication of Study~1 by testing the generalizability of the intervention effects in a different population. Hence, we followed the same experimental structure as Study~1. We used a within-subjects design to evaluate the effects of the four LLM-powered chatbot interventions, and recruited $N = 250$ participants from Kenya. Overall, 88\% of the participants were aged 18--34, 12\% aged 35--54, and almost no participants aged $55$ or older. The gender distribution was 62\% male and 38\% female (no participants identified as another gender identity). Participants indicated centrist political orientation on average ($M = 50.08$, SD = $29.74$) and, compared to the U.K. sample, participants reported substantially more frequent prior contact with refugees ($M = 45.32$, SD = $29.91$), which reflects that Kenya is among the countries hosting the largest number of refugees worldwide \cite{UNHCR.2026}.

\begin{figure}[H]
    \centering
    \vspace{-1cm}
    \includegraphics[height=0.8\textheight]{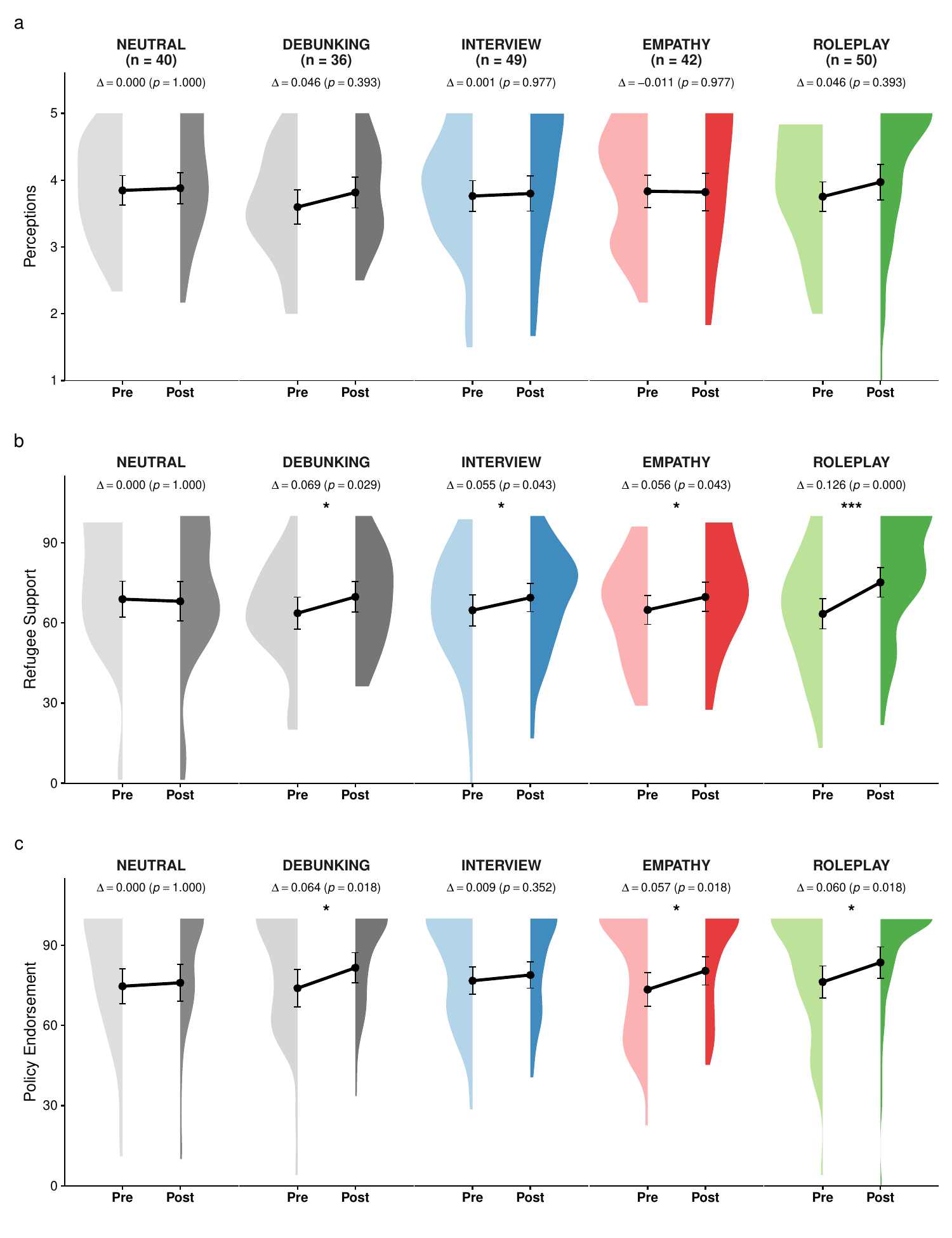}
\caption{\textbf{Study 2 ($N=250$) in Kenya to estimate the effect of LLM interactions on refugee-related outcomes.} Each panel compares the pre-and post-intervention measurements for different experimental conditions (i.e., \textsc{Neutral}, \textsc{Debunking}, \textsc{Interview}, \textsc{Empathy}, \textsc{Roleplay}). The estimated changes were analyzed using linear mixed-effects models with condition, time, and their interactions as fixed effects. The \textsc{Neutral} condition is the reference level. Violin plots show the distribution of raw measurements. The black dots show the mean, and whiskers the 95\% confidence intervals. The estimated effects ($\Delta$) represent the additional pre–post change in each intervention relative to the \textsc{Neutral} condition, where $\Delta = 0.01$ corresponds to an increase of 1 percentage point (p.p.). The $p$-values refer to Holm-corrected one-sided tests of the condition $\times$ time interaction. Significance levels: $^{\dagger} p < 0.1$, $^{*} p < 0.05$, $^{**} p < 0.01$, $^{***} p < 0.001$.}
\label{fig:Study2_Confirmatory_Stacked_Figure}
\end{figure}

\paragraph*{Perceptions of refugees.}
The LLM interventions did not affect participants' perceptions of refugees (see \Cref{fig:Study2_Confirmatory_Stacked_Figure}a). After Holm correction, no contrast differed significantly from the \textsc{Neutral} condition (all $p > 0.39$). The estimated changes were $4.6$ p.p. for \textsc{Debunking} ($b = 0.046$, SE $= 0.038$, $t = 1.202$, $p = 0.393$, 95\% CI: [$-0.029, 0.122$]), no change for \textsc{Interview} ($b = 0.001$, SE $= 0.036$, $t = 0.029$, $p = 0.977$, 95\% CI: [$-0.069, 0.071$]), $-1.1$ p.p. for \textsc{Empathy}  ($b = -0.011$, SE $= 0.037$, $t = -0.307$, $p = 0.977$, 95\% CI: [$-0.084, 0.061$]), and $4.6$ p.p. for \textsc{Roleplay} ($b = 0.046$, SE $= 0.035$, $t = 1.295$, $p = 0.393$, 95\% CI: [$-0.024, 0.116$]). Perceptions thus remained stable across conditions, consistent with Study~1. The pairwise comparison among the LLM interventions were not statistically significant.

\paragraph*{Refugee support.} The effect on refugee support varied across interventions (see \Cref{fig:Study2_Confirmatory_Stacked_Figure}b). Consistent with Study~1, the \textsc{Roleplay} intervention led to an increase in support by $12.6$ p.p., which was statistically significant after Holm correction ($b = 0.126$, SE $= 0.027$, $t = 4.653$, $p < 0.001$, 95\% CI: [$0.072, 0.179$]). The other LLM interventions also produced significant, albeit smaller, effects after Holm correction: \textsc{Debunking} led to an increase by $6.9$ p.p. ($b = 0.069$, SE $= 0.029$, $t = 2.354$, $p = 0.029$, 95\% CI: [$0.011, 0.126$]), \textsc{Interview} by $5.5$ p.p. ($b = 0.055$, SE $= 0.027$, $t = 2.034$, $p = 0.043$, 95\% CI: [$0.002, 0.109$]), and \textsc{Empathy} by $5.6$ p.p. ($b = 0.056$, SE $= 0.028$, $t = 1.997$, $p = 0.043$, 95\% CI: [$0.001, 0.112$]). As an exploratory analysis, we performed pairwise comparisons between interventions (see Supplement~\ref{suppsec:s2_pairwise}): \textsc{Roleplay} difference to \textsc{Debunking} (difference in change: $5.7$ p.p., 95\% CI: $[0.2, 11.2]$) was not significant after correction ($p = 0.170$), it improved refugee support more than \textsc{Interview} ($7.0$ p.p., 95\% CI: $[2.0, 12.1]$, $p = 0.038$), and \textsc{Empathy} ($6.9$ p.p., 95\% CI: $[1.7, 12.2]$, $p = 0.049$). \textsc{Debunking}, \textsc{Interview}, and \textsc{Empathy} did not differ from one another (all $p = 1.000$).

\paragraph*{Policy endorsement.}
Policy endorsement showed improvements for three of the four interventions (see \Cref{fig:Study2_Confirmatory_Stacked_Figure}c). The \textsc{Debunking} intervention increased policy endorsement by $6.4$ p.p. ($b = 0.064$, SE $= 0.025$, $t = 2.585$, $p = 0.018$, 95\% CI: [$0.015, 0.112$]). The \textsc{Interview} intervention produced no statistically significant change ($b = 0.009$, SE $= 0.023$, $t = 0.381$, $p = 0.352$, 95\% CI: [$-0.036, 0.054$]). The \textsc{Empathy} intervention by $5.7$ p.p. ($b = 0.057$, SE $= 0.024$, $t = 2.402$, $p = 0.018$, 95\% CI: [$0.010, 0.104$]), the \textsc{Roleplay} intervention by $6.0$ p.p. ($b = 0.060$, SE $= 0.023$, $t = 2.641$, $p = 0.018$, 95\% CI: [$0.015, 0.105$]). As an exploratory analysis with pairwise comparisons across interventions (see Supplement~\ref{suppsec:s2_pairwise}), the pairwise differences between \textsc{Debunking}, \textsc{Empathy}, and \textsc{Roleplay} were not statistically significant after Holm correction (all $p \geq 0.108$). However, \textsc{Debunking}, \textsc{Empathy}, and \textsc{Roleplay} showed larger improvements than \textsc{Interview} but were not statistically significant (\textsc{Roleplay}: $5.1$ p.p., 95\% CI: $[0.9, 9.4]$, $p = 0.108$; \textsc{Debunking}: $5.5$ p.p., 95\% CI: $[0.9, 10.1]$, $p = 0.108$; \textsc{Empathy}: $4.8$ p.p., 95\% CI: $[0.4, 9.3]$, $p = 0.135$)).

\paragraph*{Robustness checks.} Similar to Study~1, we conducted a robustness check using the \textsc{Static Information} condition instead of the \textsc{Neutral} intervention as the secondary control. Across all outcomes, the pattern of results remained consistent (see Supplement~\ref{suppsec:reanalyis_static_information}). (1)~For perceptions of refugees, we again found no improvement at the 5\%-significance level. (2)~For refugee support, the results were also robust. The \textsc{Debunking} ($b = 0.027$, $p = 0.720$), \textsc{Interview} ($b = 0.013$, $p = 0.922$), and \textsc{Empathy} ($b = 0.014$, $p = 0.922$) also showed positive shifts, though these did not reach significance. \textsc{Roleplay} intervention again showed clear improvements relative to the reference condition ($b = 0.084$, $p = 0.006$). (3)~For policy endorsement, we found a directionally consistent pattern. In particular, \textsc{Debunking} ($b = 0.043$, $p = 0.236$), \textsc{Empathy} ($b = 0.036$, $p = 0.236$), and \textsc{Roleplay} ($b = 0.039$, $p = 0.236$), showed positive estimates of similar magnitude to the main analysis, although these contrasts were no longer significant after Holm correction against the \textsc{Static Information} reference group. Taken together, these analyses corroborate the robustness of the findings.

\subsection*{Study 3 (long-term outcomes)}

Study~3 was conducted as a preregistered online experiment with $N = 554$ participants from the U.K. to assess the durability of intervention effects using a two-week follow-up survey ($N = 521$ responses) in addition to the immediate intervention effects. Unlike Studies~1 and~2, Study~3 used a between-subjects design, which makes it more challenging to detect intervention effects. The following results should therefore be interpreted in light of the expected lower statistical power. Overall 27\% of the participants were aged $18$--$34$, 46\% aged $35$--$54$, and 27\% aged $55$ or older. The gender distribution was fairly balanced (54\% female, 45\% male, and 1\% identifying as another gender identity). Participants reported moderately left-leaning political orientations ($M = 42.33$, $SD = 24.76$) and low to moderate prior contact with refugees ($M = 12.33$, $SD = 18.28$).

In terms of the immediate intervention effects, the main findings from Studies 1 and 2 were replicated (Supplement~\ref{suppsec:study_3}). For general perceptions of refugees, no intervention led to a change (all $p = 1.000$), consistent with above. For refugee support, the \textsc{Roleplay} intervention led to a significant increase ($b = 0.077$, $SE = 0.031$, $p = 0.023$, $d = 0.36$). For policy endorsement, the effect of \textsc{Roleplay} was positive and of similar size to the above studies but not statistically significant ($b = 0.076$, $SE = 0.045$, $p = 0.177$, $d = 0.25$). No other intervention showed significant differences in refugee support or policy endorsement. In the following, we focus on the two-week follow-up to assess the durability of intervention effects

\begin{figure}[H]
    \centering
    \vspace{-1cm}
    \includegraphics[height=0.9\textheight]{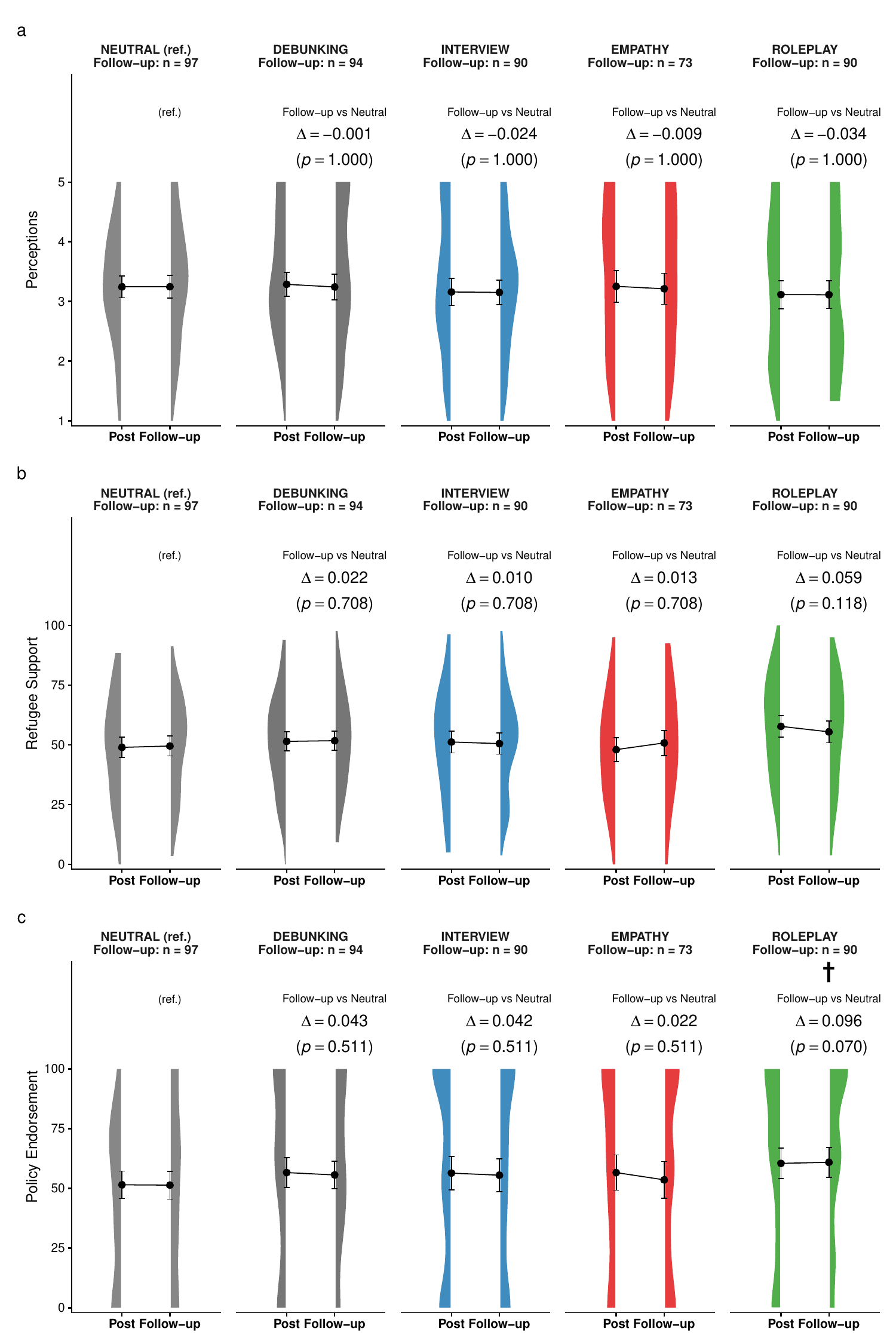}
    \caption{\textbf{Study 3 ($N = 554$; follow-up $N = 521$) in the U.K. to estimate the durability of LLM intervention effects on refugee-related outcomes.} Each panel shows the immediate outcomes (``post'') and the two-week follow-up for the five experimental conditions (i.e., \textsc{Neutral}, \textsc{Debunking}, \textsc{Interview}, \textsc{Empathy}, \textsc{Roleplay}). Following the preregistration, each outcome was analyzed with a linear regression comparing each intervention condition to the \textsc{Neutral} reference condition. The violin plots show the distribution of raw measurements for the outcomes at different time steps. The black dots show the mean, and whiskers the 95\% confidence intervals. The changes ($\Delta$) refer to the estimated difference relative to the \textsc{Neutral} condition at follow-up. The $p$-values refer to Holm-corrected one-sided tests (treatment $>$ control). Significance thresholds: $^{\dagger} < 0.1$, $^{*} p < 0.05$, $^{**} p < 0.01$, $^{***} p < 0.001$.}
    \label{fig:FollowUp_Confirmatory_Stacked_Figure}
\end{figure}

\paragraph*{Perceptions of refugees.}
Two weeks after the interaction, the LLM interventions did not affect participants' perceptions of refugees at a statistically significant level (see \Cref{fig:FollowUp_Confirmatory_Stacked_Figure}a). The estimated changes are: \textsc{Debunking} leads to a reduction by $0.1$ p.p. ($b = -0.001$, SE $= 0.038$, $t = -0.030$, 95\% CI: [$-0.076, 0.074$]), \textsc{Interview} by $2.4$ p.p. ($b = -0.024$, SE $= 0.039$, $t = -0.620$, 95\% CI: [$-0.100, 0.052$]); \textsc{Empathy} by $0.9$ p.p. ($b = -0.009$, SE $= 0.041$, $t = -0.215$, 95\% CI: [$-0.089, 0.071$]); and \textsc{Roleplay} by $3.4$ p.p. ($b = -0.034$, SE $= 0.039$, $t = -0.872$, 95\% CI: [$-0.109, 0.042$]).

\paragraph*{Refugee support.} The estimated changes are as follows (see \Cref{fig:FollowUp_Confirmatory_Stacked_Figure}b): the \textsc{Debunking} intervention increased refugee support by $2.2$ p.p. ($b = 0.022$, SE $= 0.031$, $t = 0.719$, $p = 0.708$, 95\% CI: $[-0.039, 0.083]$); \textsc{Interview} by $1.0$ p.p. ($b = 0.010$, SE $= 0.031$, $t = 0.322$, $p = 0.708$, 95\% CI: $[-0.052, 0.072]$); \textsc{Empathy} by $1.3$ p.p. ($b = 0.013$, SE $= 0.033$, $t = 0.391$, $p = 0.708$, 95\% CI: $[-0.052, 0.078]$); and \textsc{Roleplay} by $5.9$ p.p. ($b = 0.059$, SE $= 0.031$, $t = 1.895$, $p = 0.118$, 95\% CI: $[-0.002, 0.121]$, $d = 0.28$). None of these estimates were statistically significant.

\paragraph*{Policy endorsement.} The estimate effects for the follow-up are (see \Cref{fig:FollowUp_Confirmatory_Stacked_Figure}c): \textsc{Debunking} increased policy endorsement by $4.3$ p.p. ($b = 0.043$, SE $= 0.045$, $t = 0.953$, $p = 0.511$, 95\% CI: [$-0.045, 0.130$]), \textsc{Interview} by $4.2$ p.p. ($b = 0.042$, SE $= 0.045$, $t = 0.920$, $p = 0.511$, 95\% CI: [$-0.047, 0.130$]), and \textsc{Empathy} by $2.2$ p.p. ($b = 0.022$, SE $= 0.048$, $t = 0.464$, $p = 0.511$, 95\% CI: [$-0.072, 0.116$]). The \textsc{Roleplay} intervention shows a positive estimated effect of $9.6$ p.p. ($b = 0.096$, SE $= 0.045$, $t = 2.118$, $p = 0.070$, 95\% CI: [$0.007, 0.185$], $d = 0.31$), which was above the preregistered $\alpha = 0.05$ threshold after Holm correction, but was still below a more lenient $\alpha = 0.1$ threshold.

Notably, the estimated effect was similar to that observed immediately after the intervention, suggesting that the intervention could have a durable effect. Given the lower statistical power of the between-subjects design, these findings can motivate a large-scale experiment to confirm the long-term effects. Here, a sample size calculation based on the observed effect size for \textsc{Roleplay} ($d = 0.31$) suggests that approximately $200$ participants per condition (i.e., \textsc{Roleplay} and \textsc{Neutral}) would be needed to achieve $80\%$ power at a conservative one-sided significance threshold of $\alpha = 0.0125$. For comparison, an exploratory analysis without adjustment for multiple comparisons further showed positive effects of \textsc{Roleplay} on policy endorsement ($b = 0.096$, one-sided $p = 0.017$) and refugee support ($b = 0.059$, one-sided $p = 0.029$) at the two-week follow-up. The findings further suggest that the \textsc{Roleplay} intervention may have lasting effects.

\newpage
\thispagestyle{empty}

\newpage

\paragraph*{Robustness checks.}
To assess whether the follow-up results depend on the choice of reference category, we re-estimated all models using the \textsc{Static Information} instead of the \textsc{Neutral} intervention as the baseline, while keeping the model structure and outcome variables unchanged (see Supplement~\ref{suppsec:reanalyis_static_information}).

\section*{Discussion}
\label{sec:discussion}

\paragraph*{Theoretical findings.}

Across the three preregistered experiments ($N = 1358$), LLM-based interventions using \textsc{Empathy} and \textsc{Roleplay} were effective in increasing support for refugees. The effect of \textsc{Roleplay} on refugee support (Cohen's $d = 0.45$ in Study~1, $0.64$ in Study~2, and $0.36$ in the between-subjects Study~3) is comparable to or exceeds the average effect size in the prejudice-reduction literature, where the effects of intervention are often modest (i.e., $d = 0.357$) \cite{Paluck.2021}. The \textsc{Interview} intervention showed weaker and less consistent effects. Further, \textsc{Debunking} showed limited effects, which was expected, because refugee attitudes are often shaped less by factual beliefs alone than by symbolic threat, perceived deservingness, and broader social identities \cite{Bansak.2016, Hainmueller.2014, Hopkins.2019}. Interestingly, the effects were stronger for outcomes related to refugee support and policy endorsement, while general perceptions of refugees remained largely stable. This pattern is consistent with prior research showing that general evaluations of outgroups are often difficult to shift, whereas support for concrete actions and policies may be more responsive to humanitarian and perspective-taking cues \cite{Hainmueller.2014, Kalla.2023, Paluck.2021, Adida.2018}. Hence, the proposed interventions are of direct relevance for policy-makers and organizations seeking to strengthen public support for refugees.

Theoretically, the \textsc{Empathy} intervention operates through a psychological pathway centered on emotional resonance and compassionate understanding, which invites participants to connect directly with the feelings and experiences of refugees and thereby activates prosocial motivation \cite{Batson.1997, Dovidio.2004}. This may be especially effective in moralized social domains, where attitudes are often shaped by emotional and humanitarian evaluations rather than factual beliefs alone \cite{Kubin.2021, Rutland.2010}. By highlighting shared vulnerability and human experience, empathy-based dialogue may reduce identity threat \cite{Costello.2024, Sherman.2006}, increase emotional engagement \cite{Dovidio.2004}, activate humanitarian concern \cite{Bansak.2016}, and increase perceived similarity with refugees \cite{Todd.2014, Galinsky.2000}. In this way, \textsc{Empathy} can foster supportive attitudes by reducing social distance and strengthening moral concern \cite{Miklikowska.2018, Stephan.1999}.

The \textsc{Roleplay} intervention works through an immersive form of perspective-taking by asking participants to take the first-person perspective of a displaced person through a structured simulation of refugee experience. This connects to imagined intergroup contact theory, which suggests that simulated encounters can reduce prejudice when real contact is limited or politically sensitive \cite{Paluck.2021, Turner.2007}. Such imagined contact can reduce anxiety, increase empathy, and provide a psychologically safe setting for engaging with an outgroup perspective \cite{Miles.2014, West.2013, Johnson.2013}. In addition, the narrative format may reduce counterarguing by shifting attention away from abstract policy debate and toward the concrete lived experience of flight, uncertainty, and vulnerability \cite{Green.2000, Kalla.2023}. In doing so, the \textsc{Roleplay} intervention can increase psychological closeness to refugees and foster self-other overlap, both of which have been linked to lower prejudice and greater prosocial concern toward outgroups \cite{Davis.1996, Galinsky.2000, Herrera.2018}.

The limited effectiveness of \textsc{Debunking} further supports the above interpretation. Given that refugee attitudes are shaped not only by factual beliefs, but also by threat perceptions, perceived deservingness, and identity-linked values \cite{Abdelaaty.2022, Bansak.2016, Hainmueller.2014, Stephan.1999}, correcting misperceptions may have limited impact on shifting these deeper evaluative structures \cite{Hopkins.2019}. In some cases, factual correction can even heighten defensiveness by signaling ideological disagreement \cite{Costello.2024, Rabb.2025}. Together, these patterns suggest that LLM-based interventions are more effective when they move beyond correcting beliefs and instead foster emotional connection, humanitarian concern, and perspective-taking.


Finally, our study contributes to the emerging literature on LLM-based persuasion \cite{Matz.2024, Bai.2025, Costello.2024, Hackenburg.2025, Lin.2025, Xu.2025, Hoelbling.2025} but also highlights important differences. Prior work has primarily focused on shifting political preferences, such as candidate choice \cite{Argyle.2025,Lin.2025} or policy positions \cite{Argyle.2025, Bai.2025}. In contrast, our study examines attitudes toward refugees, which involved a highly moralized context and are more deeply rooted in concerns about economic burden, cultural cohesion, and perceived deservingness, and are therefore likely more resistant to change \cite{Bansak.2016, Hainmueller.2014}. This distinction is reflected in our results, where outcomes related to support and policy endorsement can be shifted through the LLM dialogue, yet while the general perceptions of refugees remain largely stable. This pattern suggests that mechanisms effective in political persuasion (e.g., informational arguments or standard messaging) may not readily transfer to refugee perception as identity-linked attitudes.

\paragraph*{Practical implications.}


Our work also has direct practical implications. First, our dialogue-based LLM interventions are lightweight and can be deployed at scale, for example, via smartphones or web-based applications, which allows broad and low-cost reach across diverse and even non-WEIRD populations. In contrast, interventions through traditional mass media campaigns, despite the broad reach, often lead to only modest effects \cite{Paluck.2021}. Our approach could thus be used by a wide range of stakeholders, including governments and non-governmental organizations (e.g., the United Nations High Commissioner for Refugees (UNHCR), UNICEF, the International Committee of the Red Cross), seeking to promote support for refugees. Second, our results provide actionable guidance for stakeholders seeking to design communication strategies around refugee support. In particular, our findings suggest that communication strategies eliciting empathy and perspective-taking should be prioritized over purely informational messages. These insights extend beyond LLM-based systems and can inform the design of broader communication formats, even in traditional media, public awareness campaigns, educational materials, serious games, or digital learning environments, where perspective-taking and immersive elements can be used to foster more supportive attitudes.


More broadly, our findings highlight that perspective-taking through narrative immersion can reduce resistance and facilitate openness in identity-linked contexts \cite{Green.2000, Kalla.2023, Paluck.2021}, which may offer a generalizable framework for designing persuasive interventions beyond the refugee domain. At the same time, using LLM-based interactions is beneficial as it offers a controlled and standardized research setting, which allows us to isolate and compare different mechanisms of attitude change in a scalable and reproducible manner that would be difficult to achieve with human-generated interventions.

\paragraph*{Strengths and limitations.}


Our study has several strengths. First, it is based on a large sample across three preregistered experiments ($N=1358$ in total), which provides evidence on the effectiveness of LLM-based interventions in the refugee context. Second, we combine samples from a Western, educated, industrialized, rich, and democratic (WEIRD) context (i.e., the United Kingdom) and a non-WEIRD context (i.e., Kenya). Thereby, our study directly follows calls to extend findings beyond Western populations \cite{Henrich.2010, Ghai.2025, Rad.2018} and explicitly tests the generalizability across contexts where refugee-related issues carry different levels of political and social relevance. The United Kingdom was chosen due to ongoing national debates about asylum policy \cite{Bansak.2016}, while Kenya provides a context in which refugee presence is highly prevalent, with more than 600,000 displaced individuals residing in refugee camps such as Dadaab and Kakuma \cite{UNHCR.2026}. Third, we used an open-weight LLM to improve the transparency and reproducibility \cite{Shrestha.2023} of our work, as the underlying model can be accessed and evaluated more easily than proprietary systems. Fourth, Study~3 included a follow-up measurement two weeks after the interaction. Longitudinal follow-ups remain rare in research on LLM-based interventions, where most studies assess attitudes only immediately after exposure \cite{noels.2026, Hoelbling.2025}. Studies that include longitudinal assessments often rely on repeated pre- and post-intervention measurements \cite{Costello.2024}, which may introduce priming, anchoring, or consistency effects. In contrast, Study~3 used a more challenging between-subjects design without baseline attitude measurements, thereby minimizing such priming or anchoring effects. Our study is thus among the first to examine the persistence of LLM-based intervention effects over time using a between-subject design.


This study has several limitations. First, participants were informed that they interacted with an LLM, which may have shaped their responses; however, this transparency was intentional for ethical reasons. Second, the interventions were implemented using a state-of-the-art open-weight LLM (i.e., gpt-oss 20b from OpenAI). The results may differ when using alternative models. Choosing more advanced, proprietary LLMs is likely to produce stronger effects but would reduce reproducibility. Third, outcomes were based on self-reported attitudes, which may be subject to social desirability bias. Still, this is a common approach in prior literature \cite{Bansak.2016, Hainmueller.2014}.


We further note that the long-term effects observed in Study~3 were weaker and did not consistently reach the preregistered significance levels, although the estimates were directionally aligned with the short-term results from Study 1 and 2. This may reflect limited statistical power in the between-subjects design, but is also consistent with prior evidence \cite{Paluck.2021, Hangartner.2021,Adida.2018} showing that brief interventions often decay over time as individuals return to their everyday social environments. Importantly, the observed patterns still suggest that LLM-based dialogues can initiate attitudinal shifts, even if our data cannot establish whether these persist after a single exposure. Future work should therefore test whether repeated conversations can reinforce such shifts and help maintain them over time. Given their scalability \cite{Breum.2023, Wang.2024}, LLM-based systems are well-suited for such repeated, personalized interactions.

\paragraph*{Broader impact.}


Our findings have broader implications amid record levels of forced displacement \cite{UnitedNationsHighCommissionerforRefugees.2025} and persistent public resistance to refugee support \cite{Dinas.2019, Hainmueller.2014}. This work highlights the potential of brief LLM-based interactions to complement traditional communication efforts. At the same time, a careful, responsible, and ethical use is mandatory, especially in sensitive societal domains to ensure transparency and prevent unintended harms.

\section*{Methods}
\label{sec:methods}

The three studies were preregistered via  \url{https://osf.io/7puzy}. All participants provided informed consent. The studies received ethical approval at LMU Munich (approval number: ETH-SOM-029). We followed the GUIDE-LLM checklist \cite{llm_checklist} for reporting LLM use in behavioral science.

\subsection*{Interventions}

We implemented four LLM-based intervention conditions (i.e., \textsc{Interview}, \textsc{Empathy}, and \textsc{Roleplay}, \textsc{Debunking}), the control condition (i.e., \textsc{Neutral}), and the secondary control (i.e., \textsc{Static Information}). All LLM interventions followed a standardized three-phase conversational architecture (the prompts are in  Supplement \Cref{suppsec:full_prompts}). Each conversation included an introduction phase, an interaction phase, and a reflection phase. In the introduction phase, the LLM established a neutral, non-judgmental tone and briefly explained the purpose of the exchange. The interaction phase contained four to twelve conversational turns, depending on the condition. During this phase, the LLM asked open-ended questions and generated responses tailored to participants' messages. The reflection phase concluded the exchange. Here, the LLM prompted participants to consider the conversation and ended with a standardized closing message. Afterwards, we asked additional survey questions (see Supplement \Cref{suppsec:survey_questions}).

The LLM interventions are as follows:

$\bullet$\,The \textsc{Debunking} intervention was structured to correct common misconceptions about refugees using evidence-based information and tailored counterarguments. Drawing on research in misinformation correction \cite{Costello.2024,Rabb.2025}, participants were asked to share specific concerns. Responses contained factual data from international sources such as UNHCR statistics. The tone remained neutral and non-confrontational to reduce defensive processing while providing clear corrective information.


$\bullet$\,The \textsc{Interview} intervention was designed to encourage participants to articulate their own values and reasoning and thus promote self-reflection, was grounded in self-persuasion theory, which suggests that people revise attitudes more effectively when they generate their own arguments \cite{Argyle.2023}. During the intervention, reflective, open-ended questions were used, and participants' statements were mirrored in a supportive tone to trigger elaboration and introspection (while new facts or counterarguments were avoided).


$\bullet$\,The \textsc{Empathy} intervention aimed to increase prosocial behavior by highlighting the emotional experiences and vulnerabilities of refugees. The rationale for this intervention builds on empathy-based interventions that have been shown to reduce negative intergroup attitudes \cite{Hangartner.2021,Baer.2024}. During the interaction phase, the participants were prompted to consider emotional challenges such as fear, loss, and uncertainty, and used empathetic language to encourage emotional engagement (rather than rational correction). 



$\bullet$\,The \textsc{Roleplay} intervention was designed to guide participants through a first-person narrative to induce perspective-taking through an immersive simulation \cite{Kalla.2023}. Participants were placed in scenarios such as leaving home or arriving at a border crossing, and were asked to describe their thoughts and emotions within these situations.


$\bullet$\,The \textsc{Neutral} intervention followed prior literature \cite{Xu.2025} and included a short non-political conversation about preferences between cats and dogs.

$\bullet$\,The \textsc{Static Information} intervention in form of a flyer is a condition, providing a concise, neutrally phrased written summary of refugee-related facts (the content is in Supplement~\ref{suppsec:control_condition}). The general design followed recent studies about LLM-powered persuasion \cite{Xu.2025}. We later use this condition as a secondary reference category to isolate the effect of the interactive dialogue. 

The prompts for the above interventions were developed iteratively with feedback from individuals experienced in supporting and interacting with refugees through volunteer work at non-governmental organizations (e.g., UNICEF). We conducted several small pilot runs to collect qualitative feedback and then to refine the wording, structure, and conversational flow. We also screened the generated dialogues for potential safety issues, inconsistencies, or unintended model behaviors. No safety-relevant or otherwise problematic patterns were observed during the pilot or in the main experiment.

The number of conversational rounds was specified in the intervention prompts and varied according to the underlying psychological mechanism. The \textsc{Interview}, \textsc{Empathy}, and \textsc{Debunking} interventions each comprised six rounds, whereas \textsc{Roleplay} included twelve rounds to allow sufficient time for narrative immersion across six scenarios. The \textsc{Neutral} condition was limited to four rounds to minimize participant fatigue. To account for potential differences in length and thus engagement, we conducted a robustness check; after controlling for duration of conversations, the main findings remained largely consistent, with \textsc{Roleplay} showing significant improvements in refugee support (see Supplement~\ref{suppsec:engagementControl}).

\paragraph*{Implementation details.} 

The interactive LLM conditions were delivered through a custom Python application built with the Streamlit web framework (https://streamlit.io). We used a frontier, open-weight LLM (namely, gpt-oss 20b from OpenAI \cite{openai_gpt_oss_20b}), which we accessed via the Together.ai API in real time. The temperature was set to $0.7$. This allowed natural, yet instruction-consistent responses. All other parameters remained at default values.

\subsection*{Measures}

The effect of the interventions was assessed using three dimensions (see Supplement~\ref{suppsec:survey_questions}):
\begin{enumerate}
\item \textit{Perceptions of refugees:} This was measured using a composite index adapted from \cite{Hangartner.2019}, where participants indicated their agreement with statements assessing perceived warmth, trustworthiness, and general impressions of refugees. Responses were recorded on a 5-point Likert scale (1 = \textit{completely disagree}, 5 = \textit{completely agree}). Higher scores reflected more positive perceptions. 
\item \textit{Refugee support:} The general tendency of compassion, moral obligation, and the willingness to provide personal or community support was measured using items adapted from \cite{Czarnek.2025b}. Participants responded using a slider from 0 to 100. Higher values indicated stronger prosocial motivation and greater perceived responsibility.
\item \textit{Policy endorsement:} Support for refugee-related policies was measured using items adapted from \cite{Vlasceanu.2024}. Participants rated their agreement with policy proposals such as asylum procedures, access to social services, and institutional support. Responses were recorded using a slider from 0 to 100 (0 = \textit{completely disagree}, 100 = \textit{completely agree}). Higher scores indicated stronger endorsement of refugee-supportive policies.
\end{enumerate}
Composite indices were constructed by averaging the respective items. Internal consistency was evaluated using Cronbach's alpha. Higher scores consistently indicated more positive attitudes toward refugees. The same scoring procedure was applied at different measurement points to ensure comparability across time.

The measures were collected via Qualtrics. At the beginning of the survey, we additionally recorded sociodemographics (e.g., age, gender, etc.), as well as political orientation and prior contact with refugees. The exact questions are in Supplement~\ref{suppsec:survey_questions}.

\subsection*{Study 1}
\subsubsection*{Participants}

Study 1 was conducted on March~31, 2026 in the U.K. The U.K. was chosen due to ongoing national debates about asylum policy \cite{Bansak.2016}. Participants were recruited through Prolific (https://www.prolific.com) and were compensated with GBP~9. Eligibility criteria required participants (1)~to be at least $18$ years old, (2)~report English as their first language, (3)~hold a Prolific approval rating of 98\% or higher, and (4)~have completed at least $50$ previous studies. All participants gave informed consent.

An a priori power analysis was conducted in G*Power (version 3.1.9.7) using a $t$-test for dependent means. The analysis assumed an effect size of $d = 0.35$, a Holm-corrected significance level of $0.0125$, and 80\% power, which indicated a minimum of $486$ participants (i.e., $81$ per group). To account for attrition, the target sample was increased by 20\%. This resulted in an intended sample size of $N = 583$. After applying the preregistered exclusion criteria (in Supplement~\ref{suppsec:inclusion_exclusion}), the final sample consisted of $N = 554$ participants, with roughly balanced group sizes across the six conditions. 
 
\subsubsection*{Procedure}

Study 1 used a randomized experimental design with a within-subjects pre--post comparison. Random assignment was implemented using the built-in randomization function in Qualtrics.

The experimental procedure included three phases: (1)~a pre-survey, (2)~the randomized intervention, and (3)~an immediate post-survey. During the pre-survey, participants provided informed consent and completed demographic questions, including age, gender, nationality, education level, political orientation, and prior contact with refugees (see Supplement~\ref{suppsec:survey_questions}). As part of the onboarding, participants were explicitly informed that the conversation partner in the intervention phase was an LLM. The post-survey also included an attention check and an open-ended reflection item, and a debriefing.

\subsubsection*{Analysis}


Changes from pre- to post-measurement were analyzed using linear mixed-effects models. Each dependent variable was modeled separately. Independent variables were time (pre vs. post) and condition (five intervention groups; \textsc{Static Information} excluded) as fixed effects. We further included an interaction Condition $\times$ Time to test whether the pre--post change differed across interventions. The \textsc{Neutral} condition served as the reference category. The confirmatory tests from our preregistration focused on the interaction contrasts comparing the pre--post change in each intervention against the \textsc{Neutral} group. All $p$-values were adjusted using Holm correction. We preregistered a significance level of $\alpha = 0.05$. We repeated the analysis using \textsc{Static Information} as the secondary control in Supplements~\ref{suppsec:reanalyis_static_information}. For estimation, we used the \texttt{lme4} package in R (version 4.5.2).


In exploratory analyses, we also assessed the comparative effectiveness of the interventions. To do so, we tested pairwise contrasts of the Condition $\times$ Time interaction terms in the linear mixed-effects models. These comparisons were tested two-sided, as no directional hypotheses were preregistered. We again applied Holm correction.


\subsection*{Study 2}

Study~2 was conducted between April~8, 2026 and April ~22, 2026 and followed the design of Study~1, with the exception that the study was conducted in Kenya. Kenya hosts one of the largest refugee populations in Africa and faces ongoing public debates about migration and humanitarian policy \cite{UnitedNationsHighCommissionerforRefugees.2025}. Participants were recruited again through Prolific, but the eligibility criteria were broadened slightly to ensure a sufficient recruitment pool: (1)~minimum age of $18$, (2)~fluency in English, and (3)~an approval rating of at least 98\%. Compensation was GBP~9. All participants gave informed consent.


Following the power analysis from Study 1, we again aimed to recruit $N = 583$ participants. However, the target sample size was not reached within two weeks, which reflects that the available participant pool on Prolific was smaller for Kenya than for the U.K. Hence, different from the preregistration, the data collection was stopped after weeks, after which $271$ participants had completed the survey. After exclusions, the final sample consisted of $N = 250$ participants.

All other parts of the study (i.e., interventions, measures, procedure, and analysis) were identical to Study~1.

\subsection*{Study 3}

\subsubsection*{Participants}


Study 3 used a between-subjects design and was conducted on January~23, 2026, with a follow-up on February 6, 2026. Participants were recruited through Prolific from the U.K. The same eligibility criteria as in Study 1 were applied. Compensation was GBP~9 for the initial study and GBP~9 for the follow-up. All participants provided informed consent.

Following our preregistration, we aimed to recruit $N = 562$ participants for Study~3. In total, $N = 554$ participants completed the main study, out of which $N = 521$ completed the follow-up survey. The random assignment produced roughly balanced group sizes across conditions. The demographic characteristics were similar to Study 1, with slight differences in political orientation and prior contact with refugees.

\subsubsection*{Procedure}


Study 3 followed a longitudinal design with three measurement points: a pre-survey, a post-survey, and a two-week follow-up. The pre-survey collected demographic information only. Both the post-survey and the follow-up collected our main measures (i.e., perceptions of refugees, refugee support, policy endorsement). These measures were not included in the pre-survey to avoid priming participants or anchoring their later responses. The detailed questions are in Supplement~\ref{suppsec:survey_questions}. The procedure was otherwise identical to Studies~1 and 2.

\subsubsection*{Analysis}

To assess the immediate intervention effects, we fitted ordinary least-squares regression models for each outcome, with condition as the predictor and \textsc{Neutral} as the reference category (i.e., excluding \textsc{Static Information}). Following our preregistration, we tested the four intervention contrasts against \textsc{Neutral} using one-sided tests with Holm correction ($\alpha = 0.05$). 

To assess the durability of the intervention effects, we fitted linear mixed-effects models for each outcome, where we include a Condition $\times$ Time interaction to capture changes in the intervention effects relative to \textsc{Neutral} between the two measurements. We tested the effect by one-sided tests of the four intervention-specific changes with Holm correction (Supplement~\ref{suppsec:study_3}). In exploratory analyses, we also compared the interventions with one another at both measurement points using two-sided pairwise contrasts with Holm correction across the six intervention pairs within each outcome (Supplement~\ref{suppsec:study_3}). We used the packages \texttt{lme4} and \texttt{lmerTest} in R.

\subsection*{Robustness checks}

We ran different robustness checks in our study. First, we report Cronbach's alpha to assess internal consistency between our dependent variables (i.e., perceptions, refugee support and policy endorsement) across studies (see Supplement~\ref{suppsec:reliability}). 

Second, we use the condition \textsc{Static Information} as a reference category to test whether our analysis is sensitive to the choice of our control group. This allowed us to assess whether the intervention effects remained consistent when compared against a non-interactive, information-based intervention rather than a neutral chatbot. The results are reported in Supplement~\ref{suppsec:reanalyis_static_information} across three studies.

Third, we assessed whether the intervention effects remained robust after controlling for chat duration, measured by the time participants spent interacting with the chatbot (Supplement~\ref{suppsec:engagementControl}).

Fourth, to understand how participants experienced the dialogues, we asked them in a post-survey to rate the chatbot interactions on perceived empathy, pleasure, and self-reflection (0--100 sliders; see Supplement~\ref{suppsec:s1_Manipulation_check}). As a result, we generally find that the LLM interventions elicit the underlying psychological mechanism (e.g., the \textsc{Empathy} intervention is perceived more positively on the empathy scale than the control).

\vspace{0.4cm}
\section*{Data availability}

\vspace{0.4cm}

All code to replicate our analyses is available via anonymous repository at \url{https://github.com/vosh-96/Refugee_support_chatbot_intervention}.


\newpage
\bibliography{literature}


\newpage
\section*{Acknowledgments}

Funding by the German Research Foundation (Grant: 543018872) is acknowledged.

\vspace{0.4cm}
\section*{Author contributions} 

K.B. performed the experiment and analyzed the data. K.B. prepared the first draft, with the help of S.F. A.B. supported in the analysis and the manuscript writing. All authors contributed to conceptualization, manuscript editing, and approved the manuscript.

\vspace{0.4cm}
\section*{Competing interests}
The authors declare no competing interests.

\newpage

\appendix

\setcounter{table}{0}
\setcounter{figure}{0}
\renewcommand{\thetable}{S\arabic{table}}
\renewcommand{\thefigure}{S\arabic{figure}}
\renewcommand{\tablename}{Supplementary Table}
\renewcommand{\figurename}{Supplementary Figure}

\begin{center}
\LARGE \textbf{Supplementary Materials}
\end{center}
\vspace{1cm}

\clearpage

\section{Participant inclusion/exclusion}
\label{suppsec:inclusion_exclusion}

\begin{table}[H]
\centering
\footnotesize
\caption{\textbf{Participant inclusion/exclusion}. The preregistered exclusion criteria were identical across the studies (i.e., a failed attention check; an unreasonably fast completion time, defined as a completion of under one-third of the condition-specific median time; or fewer than three words in the mandatory reflection item to exclude people with low effort). 
}
\scriptsize
\begin{tabular}{lccc}
\toprule
 & \textbf{Study 1 (U.K.)} & \textbf{Study 2 (Kenya)} & \textbf{Study 3 (U.K.)} \\
Design & Within-subjects & Within-subjects & Between-subjects \\
\midrule
Complete response & 583 & 271 & 562 \\
After removing duplicates/incomplete & 563 & 260 & 559 \\
After removing submissions returned on Prolific & 563 & 260 & 558 \\
\midrule
\#Excluded due to failed attention checks & 0 & 9 & 0 \\
\#Excluded due to unreasonably fast completion time & 0 & 1 & 1 \\
\#Excluded due to low effort & 9 & 0 & 3 \\
\hline
\textbf{Final sample size} & \textbf{554} & \textbf{250} & \textbf{554} \\
\hline
\hline
Completed follow-up & --- & --- & 522 \\
\midrule
\#Excluded due to failed attention checks & --- & --- & 0 \\
\#Excluded due to unreasonably fast completion time & --- & --- & 0 \\
\#Excluded due to low effort & --- & --- & 1 \\
\hline
\textbf{Final follow-up sample} & --- & --- & \textbf{521} \\
\bottomrule
\end{tabular}
\label{tab:si_samples}
\end{table}

\clearpage

\section{Sociodemographic characteristics}
\label{suppsec:demographics}

Here, we report summary statistics for the sociodemographic variables for Study~1, for Study~2, and for Study~3, all after applying the exclusion criteria.

\begin{table}[H]
\centering
\singlespacing
\begin{table}
\centering
\caption{\textbf{Sociodemographic characteristics of Study~1} (categorical variables).}
\scriptsize
\begin{tabular}[t]{llrr}
\hline
Variable & Category & $n$ & Percent\\
\midrule
Age & Young & 214 & 38.6\\
 & Middle & 246 & 44.4\\
 & Older & 94 & 17.0\\
 & Unknown & 0 & 0.0\\
\midrule
Gender & Male & 248 & 44.8\\
 & Female & 304 & 54.9\\
 & Other & 2 & 0.4\\
\midrule
Employment & Full-time & 305 & 55.1\\
 & Part-time/Student & 97 & 17.5\\
 & Other/Not working & 152 & 27.4\\
\midrule
Children & No children & 287 & 51.8\\
 & Has children & 267 & 48.2\\
\midrule
Economic situation & Low income & 177 & 31.9\\
 & Middle income & 342 & 61.7\\
 & High income & 24 & 4.3\\
 & Prefer not to say & 11 & 2.0\\
\midrule
Religion & None & 323 & 58.3\\
 & Christian & 183 & 33.0\\
 & Other & 48 & 8.7\\
\midrule
Migration background & Yes & 74 & 13.4\\
 & No & 475 & 85.7\\
 & Prefer not to say & 5 & 0.9\\
\bottomrule
\end{tabular}
\end{table}

\label{tab:study1_demographics_categorical}
\end{table}


\begin{table}[H]
\centering
\singlespacing
\begin{table}
\centering
\caption{\textbf{Sociodemographic characteristics of Study~1} (continuous variables).}
\scriptsize
\begin{tabular}[t]{lrr}
\toprule
Variable & M & SD\\
\midrule
Contact with refugees (0 = Never, 100 = Very often) & 13.36 & 18.23\\
Political views on economic issues (0 = Very liberal, 100 = Very conservative) & 44.54 & 25.63\\
\bottomrule
\end{tabular}
\end{table}

\label{tab:study1_demographics_continuous}
\end{table}

\clearpage

\begin{table}[H]
\centering
\singlespacing
\begin{table}
\centering
\caption{\textbf{Sociodemographic characteristics of Study~2} (categorical variables).}
\scriptsize
\centering
\begin{tabular}[t]{llrr}
\toprule
Variable & Category & $n$ & Percent\\
\midrule
Age & Young & 220 & 88.0\\
 & Middle & 29 & 11.6\\
 & Older & 1 & 0.4\\
 & Unknown & 0 & 0.0\\
\midrule
Gender & Male & 155 & 62.0\\
 & Female & 95 & 38.0\\
 & Other & 0 & 0.0\\
\midrule
Employment & Full-time & 95 & 38.0\\
 & Part-time/Student & 96 & 38.4\\
 & Other/Not working & 59 & 23.6\\
\midrule
\addlinespace
Children & No children & 111 & 44.4\\
 & Has children & 139 & 55.6\\
\midrule
Economic situation & Low income & 66 & 26.4\\
 & Middle income & 177 & 70.8\\
 & High income & 4 & 1.6\\
 & Prefer not to say & 3 & 1.2\\
\midrule
Religion & None & 11 & 4.4\\
 & Christian & 228 & 91.2\\
 & Other & 11 & 4.4\\
\midrule
Migration background & Yes & 37 & 14.8\\
 & No & 208 & 83.2\\
 & Prefer not to say & 5 & 2.0\\
\bottomrule
\end{tabular}
\end{table}

\label{tab:study2_demographics_categorical}
\end{table}


\begin{table}[H]
\centering
\singlespacing
\begin{table}
\centering
\caption{\textbf{Sociodemographic characteristics of Study~2} (continuous variables).}
\scriptsize
\centering
\begin{tabular}[t]{lrr}
\toprule
Variable & M & SD\\
\midrule
Contact with refugees (0 = Never, 100 = Very often) & 45.32 & 29.91\\
Political views on economic issues (0 = Very liberal, 100 = Very conservative) & 50.08 & 29.74\\
\bottomrule
\end{tabular}
\end{table}

\label{tab:study2_demographics_continuous}
\end{table}

\clearpage

\begin{table}[H]
\centering
\singlespacing
\begin{table}
\centering
\caption{\label{tab:s3-demo-cat}\textbf{Sociodemographic characteristics in Study 3} (categorical variables). 
}
\scriptsize
\centering
\begin{tabular}[t]{llrr}
\toprule
Variable & Category & $n$ & Percent\\
\midrule
Age & Young & 147 & 26.5\\
 & Middle & 257 & 46.4\\
 & Older & 150 & 27.1\\
 & Unknown & 0 & 0.0\\
\midrule
Gender & Male & 249 & 44.9\\
 & Female & 300 & 54.2\\
 & Other & 5 & 0.9\\
\midrule
Employment & Full-time & 268 & 48.4\\
 & Part-time/Student & 105 & 19.0\\
 & Other/Not working & 181 & 32.7\\
\midrule
Children & No children & 231 & 41.7\\
 & Has children & 323 & 58.3\\
\midrule
Economic situation & Low income & 179 & 32.3\\
 & Middle income & 341 & 61.6\\
 & High income & 22 & 4.0\\
 & Prefer not to say & 12 & 2.2\\
\midrule
Religion & None & 330 & 59.6\\
 & Christian & 195 & 35.2\\
 & Other & 29 & 5.2\\
\midrule
Migration background & Yes & 57 & 10.3\\
 & No & 490 & 88.4\\
 & Prefer not to say & 7 & 1.3\\
\bottomrule
\end{tabular}
\end{table}

\label{tab:study3_demographics_categorical}
\end{table}


\begin{table}[H]
\centering
\singlespacing
\begin{table}
\centering
\caption{\label{tab:s3-demo-cont}\textbf{Sociodemographics in Study 3} (continuous variables).}
\centering
\scriptsize
\begin{tabular}[t]{lrr}
\toprule
Variable & M & SD\\
\midrule
Contact with refugees (0 = Never, 100 = Very often) & 12.33 & 18.28\\
Political views on economic issues (0 = Very liberal, 100 = Very conservative) & 42.33 & 24.76\\
\bottomrule
\end{tabular}
\end{table}

\label{tab:study3_demographics_continuous}
\end{table}

\clearpage

\section{Reliability of measures}
\label{suppsec:reliability}

Each measure was computed as the mean of the corresponding items. We report Cronbach's alpha to assess internal consistency, which was satisfactory to excellent across studies. We report the corresponding values in Supplementary~Table~\ref{tab:study1_reliability_summary_table} (Study~1), Supplementary~Table~\ref{tab:study2_reliability_summary} (Study~2), Supplementary~Table~\ref{tab:study3_reliability_summary_post} (Study~3; immediate post-intervention), and Supplementary~Table~\ref{tab:study3_reliability_summary_followup} (Study~3; follow-up).

\begin{table}[H]
\centering
\begin{table}
\centering
\caption{\textbf{Reliability for the dependent variables (i.e., perceptions, refugee support, and
policy endorsement) in Study~1.}}
\label{tab:study1_reliability_summary_table}
\scriptsize
\centering
\begin{tabular}[t]{lr}
\toprule
Timepoint & Cronbach's alpha\\
\midrule
\addlinespace[0.3em]
\hline
\multicolumn{2}{l}{\textbf{Perceptions}}\\
\hspace{1em}Pre & 0.855\\
\hspace{1em}Post & 0.880\\
\addlinespace[0.3em]
\hline
\multicolumn{2}{l}{\textbf{Refugee support}}\\
\hspace{1em}Pre & 0.779\\
\hspace{1em}Post & 0.838\\
\addlinespace[0.3em]
\hline
\multicolumn{2}{l}{\textbf{Policy endorsement}}\\
\hspace{1em}Pre & 0.959\\
\hspace{1em}Post & 0.968\\
\bottomrule
\end{tabular}
\end{table}

\end{table}

\begin{table}[H]
\centering
\begin{table}
\centering
\caption{\textbf{Reliability summary for the dependent variables (i.e., perceptions, refugee support, and
policy endorsement) in Study~2.}}
\label{tab:study2_reliability_summary}
\scriptsize
\centering
\begin{tabular}[t]{lr}
\toprule
Timepoint & Cronbach's alpha\\
\midrule
\addlinespace[0.3em]
\hline
\multicolumn{2}{l}{\textbf{Perceptions}}\\
\hspace{1em}Pre & 0.722\\
\hspace{1em}Post & 0.790\\
\addlinespace[0.3em]
\hline
\multicolumn{2}{l}{\textbf{Refugee support}}\\
\hspace{1em}Pre & 0.720\\
\hspace{1em}Post & 0.714\\
\addlinespace[0.3em]
\hline
\multicolumn{2}{l}{\textbf{Policy endorsement}}\\
\hspace{1em}Pre & 0.895\\
\hspace{1em}Post & 0.908\\
\bottomrule
\end{tabular}
\end{table}

\end{table}

\begin{table}[H]
\centering
\begin{table}
\centering
\caption{\label{tab:study3_reliability_summary_post}\textbf{Reliability of the immediate post-intervention outcomes in Study~3.}}
\centering
\scriptsize
\begin{tabular}[t]{lr}
\toprule
Scale & Cronbach's alpha\\
\hline\hline
Perceptions (immediate post-survey) & 0.847\\
\midrule
Refugee support (immediate post-survey) & 0.750\\
\midrule
Policy endorsement (immediate post-survey) & 0.962\\
\bottomrule
\end{tabular}
\end{table}

\end{table}
\clearpage

\begin{table}[H]
\centering
\begin{table}
\centering
\caption{\textbf{Reliability summary for the follow-up outcomes (i.e., perceptions, refugee support, and policy endorsement) in Study~3.}}
\label{tab:study3_reliability_summary_followup}
\scriptsize
\centering
\begin{tabular}[t]{lr}
\toprule
Scale & Cronbach's alpha\\
\midrule
\midrule
Perceptions (follow-up) & 0.837\\
\midrule
Refugee support (follow-up) & 0.778\\
\midrule
Policy endorsement (follow-up) & 0.958\\
\bottomrule
\end{tabular}
\end{table}

\end{table}
\clearpage

\section{Model specification}
\label{suppsec:models}

Here, we report the full specification of the statistical models underlying the results in the main paper. All outcomes (i.e., perceptions of refugees, refugee support, and policy endorsement) were normalized to the 0--1 interval and modeled separately. Throughout, we use the index $i = 1, \ldots, N$ to refer to participants, and $g \in \mathcal{G} = \{\textsc{Debunking}, \textsc{Interview}, \textsc{Empathy}, \textsc{Roleplay}\}$ refers to the four LLM-based interventions. The \textsc{Neutral} condition is the omitted reference category, so that all reported coefficients are relative to \textsc{Neutral} condition. The \textsc{Static Information} condition is excluded from the main models and replaces the \textsc{Neutral} condition in the robustness checks (see Supplement~\ref{suppsec:reanalyis_static_information}).

\paragraph*{Studies 1 and 2 (within-subjects).}

In Studies~1 and~2, each participant is observed twice, before and after the intervention. Let $t \in \{0,1\}$ index the measurement time point, where $t = 0$ denotes the pre-intervention survey and $t = 1$ the post-intervention survey, and let $Y_{it}$ denote the outcome of participant~$i$ at time~$t$. We define two sets of indicator variables: $\text{Time}_t = 1$ if $t = 1$ and $0$ otherwise; and $\text{Group}_{ig} = 1$ if participant~$i$ was randomized to condition~$g$ and $0$ otherwise. For each outcome, we then fitted the following linear mixed-effects model:
\begin{equation*}
Y_{it} = \beta_0 + \beta_1\,\text{Time}_t + \sum_{g \in \mathcal{G}}\gamma_g\,\text{Group}_{ig} + \sum_{g \in \mathcal{G}}\delta_g\,\big(\text{Group}_{ig}\times\text{Time}_t\big) + u_i + \varepsilon_{it} .
\end{equation*}
Here, $u_i \sim \mathcal{N}(0,\sigma_u^2)$ is a participant-specific random intercept that captures between-person differences in the outcome, and $\varepsilon_{it} \sim \mathcal{N}(0,\sigma_\varepsilon^2)$ is the residual error. The coefficients are interpreted as follows: $\beta_0$ is the expected pre-intervention outcome in the \textsc{Neutral} condition; $\beta_1$ is the pre--post change in the \textsc{Neutral} condition; $\gamma_g$ is the baseline (pre-intervention) difference between condition~$g$ and \textsc{Neutral}, which is expected to be zero under successful randomization; and $\delta_g$ is the difference between the pre--post change in condition~$g$ and the pre-post change in \textsc{Neutral}.

The preregistered treatment effects are the four interaction terms $\delta_g$. These are the quantities reported as $\Delta$ in \Cref{fig:Confirmatory_Stacked_Figure} and \Cref{fig:Study2_Confirmatory_Stacked_Figure}. Each $\delta_g$ was tested using one-sided in the preregistered direction (intervention $>$ \textsc{Neutral}) using $\alpha = 0.05$, and the resulting four $p$-values were adjusted for multiple comparisons using the Holm procedure.

\paragraph*{Study 3 (between-subjects with long-term outcomes).}

In Study~3, the outcome measures were not collected before the intervention in order to avoid priming participants. Treatment effects are therefore identified between subjects rather than from pre-post change. Study~3 has two post-treatment measurement time points, namely, the immediate post-intervention survey and the two-week follow-up survey.

(1)~We estimated the treatment effects separately at each measurement time point. For each outcome, we fitted an ordinary least squares (OLS) regression with condition as the only predictor and \textsc{Neutral} as the reference category:
\begin{equation*}
Y_i = \beta_0 + \sum_{g \in \mathcal{G}}\beta_g\,\text{Group}_{ig} + \varepsilon_i ,
\end{equation*}
where $\varepsilon_i \sim \mathcal{N}(0,\sigma_\varepsilon^2)$. Here, $\beta_g$ is the difference in the expected outcome between condition~$g$ and \textsc{Neutral}, and thus directly gives the treatment effect of intervention~$g$. This model was estimated twice (i.e., once for the immediate post-intervention responses and once for the two-week follow-up responses). We then test the four contrasts $\beta_g$ using one-sided tests (intervention $>$ \textsc{Neutral}) with $\alpha = 0.05$ and Holm-correction. 

(2)~To assess the durability of the intervention effects, we tested whether the decay from the immediate post-intervention measurement to the follow-up was smaller in each intervention than in \textsc{Neutral}. We fitted:
\begin{equation*}
  Y_{it} = \beta_0 + \sum_{g} \gamma_g\, \mathrm{Group}_{ig}
         + \beta_1\, \mathrm{FollowUp}_t
         + \sum_{g} \delta_g\, \mathrm{Group}_{ig} \times \mathrm{FollowUp}_t
         + u_i + \varepsilon_{it}
  \label{eq:study3-durability}
\end{equation*}
for each outcome, where $g$ indexes the four interventions, $\mathrm{FollowUp}_t$ is 0 at the immediate post-intervention measurement and 1 at the two-week follow-up, and $u_i \sim N(0, \sigma_u^2)$ is a participant random intercept. Here, $\gamma_g$ is the treatment-vs-\textsc{Neutral} difference at the immediate post-intervention measurement; $\beta_1$ is the post-to-follow-up change in the \textsc{Neutral} condition; and $\delta_g$ is the change in the treatment-vs-\textsc{Neutral} difference from post to follow-up.

\paragraph*{Estimation.}

All models were estimated in R (version 4.5.2). The linear mixed-effects models were fitted using the \texttt{lme4} package, the OLS models using the base \texttt{lm()} function. The contrasts and estimated marginal means were computed using the \texttt{emmeans} package.

\newpage

\section{Re-analysis using \textsc{Static Information} as the reference condition}
\label{suppsec:reanalyis_static_information}

As a robustness check, we re-estimated the models across all three studies using \textsc{Static Information} instead of \textsc{Neutral} as the reference condition. This comparison provides a more demanding benchmark because the static information condition already presents participants with evidence-based arguments about refugees, which may themselves promote more supportive attitudes. We therefore expect smaller estimated intervention effects relative to this baseline. Nevertheless, the results remain largely consistent with our main findings. In particular, \textsc{Roleplay} continues to produce statistically significant improvements in refugee support in both Study~1 ($b = 0.075$, $p < 0.001$) and Study~2 ($b = 0.084$, $p = 0.006$), even when compared against the \textsc{Static Information} condition. For policy endorsement, the estimates remain positive but are generally not statistically significant. Overall, these findings suggest that the benefits of immersive, dialogue-based perspective-taking over traditional, fact-based arguments.

\subsection{Study~1}
\label{suppsec:roubstnessCheck_study1}

\begin{table}
\centering
\caption{\label{tab:s1-dv1-static} \textbf{Re-analysis  using \textsc{Static Information} as reference condition in Study~1 (perceptions).} Estimation result for linear mixed-effects model with a participant random intercept. Only the five Condition $\times$ Post interaction terms are reported; one-sided tests with Holm-correction. Significance levels: $^{\dagger}p<0.1$, $^{*}p<0.05$, $^{**}p<0.01$, $^{***}p<0.001$.}

\centering
\scriptsize
\begin{tabular}[t]{llllllll}
\toprule
& Estimate & SE & $t$ & LB$_{95}$ & UB$_{95}$ & $p$-value & Sig.\\
\midrule
\textsc{Neutral} × Post & $-$0.014 & 0.022 & $-$0.667 & $-$0.057 & 0.028 & 1.000 & \\
\textsc{Debunking} × Post & $-$0.018 & 0.022 & $-$0.818 & $-$0.062 & 0.026 & 1.000 & \\
\textsc{Interview} × Post & 0.001 & 0.023 & 0.065 & $$-$$0.043 & 0.046 & 1.000 & \\
\addlinespace
\textsc{Empathy} × Post & 0.027 & 0.022 & 1.219 & $-$0.016 & 0.070 & 0.558 & \\
\textsc{Roleplay} × Post & 0.025 & 0.023 & 1.095 & $-$0.020 & 0.070 & 0.558 & \\
\bottomrule
\end{tabular}
\end{table}

\begin{table}
\centering
\caption{\label{tab:s1-dv2-static}\textbf{Re-analysis  using \textsc{Static Information} as reference condition in Study~1 (refugee support).} Estimation result for linear mixed-effects model with a participant random intercept. Only the five Condition $\times$ Post interaction terms are reported; one-sided tests with Holm-correction. Significance levels: $^{\dagger}p<0.1$, $^{*}p<0.05$, $^{**}p<0.01$, $^{***}p<0.001$.}
\centering
\scriptsize
\begin{tabular}[t]{llllllll}
\toprule
 & Estimate & SE & $t$ & LB$_{95}$ & UB$_{95}$ & $p$-value & Sig.\\
\midrule
\textsc{Neutral} × Post & $-$0.022 & 0.014 & $-$1.555 & $-$0.049 & 0.006 & 1.000 & \\
\textsc{Debunking} × Post & $-$0.011 & 0.014 & $-$0.771 & $-$0.039 & 0.017 & 1.000 & \\
\textsc{Interview} × Post & 0.008 & 0.015 & 0.527 & $-$0.021 & 0.037 & 0.898 & \\
\addlinespace
\textsc{Empathy} × Post & 0.015 & 0.014 & 1.033 & $-$0.013 & 0.042 & 0.604 & \\
\textsc{Roleplay} × Post & 0.075 & 0.015 & 5.173 & 0.047 & 0.104 & $<$0.001 & ***\\
\bottomrule
\end{tabular}
\end{table}

\begin{table}
\centering
\caption{\label{tab:s1-dv3-static}\textbf{Re-analysis  using \textsc{Static Information} as reference condition in Study~1 (policy endorsement).} Estimation result for linear mixed-effects model with a participant random intercept. Only the five Condition $\times$ Post interaction terms are reported; one-sided tests with Holm-correction. Significance levels: $^{\dagger}p<0.1$, $^{*}p<0.05$, $^{**}p<0.01$, $^{***}p<0.001$.}
\centering
\scriptsize
\begin{tabular}[t]{llllllll}
\toprule
 & Estimate & SE & $t$ & LB$_{95}$ & UB$_{95}$ & $p$-value & Sig.\\
\midrule
\textsc{Neutral} × Post & $-$0.030 & 0.011 & $-$2.594 & $-$0.052 & $-$0.007 & 1.000 & \\
\textsc{Debunking} × Post & $-$0.017 & 0.012 & $-$1.427 & $-$0.040 & 0.006 & 1.000 & \\
\textsc{Interview} × Post & $-$0.020 & 0.012 & $-$1.658 & $-$0.044 & 0.004 & 1.000 & \\
\addlinespace
\textsc{Empathy} × Post & $-$0.005 & 0.012 & $-$0.457 & $-$0.028 & 0.017 & 1.000 & \\
\textsc{Roleplay} × Post & 0.028 & 0.012 & 2.294 & 0.004 & 0.051 & 0.055 & $\dagger$\\
\bottomrule
\end{tabular}
\end{table}

\newpage

\subsection{Study~2}
\label{suppsec:roubstnessCheck_study2}

\begin{table}
\centering
\caption{\label{tab:s2-dv1-static}\textbf{Re-analysis  using \textsc{Static Information} as reference condition in Study~2 (perceptions).} Estimation result for linear mixed-effects model with a participant random intercept. Only the five Condition $\times$ Post interaction terms are reported; one-sided tests with Holm-correction. Significance levels: $^{\dagger}p<0.1$, $^{*}p<0.05$, $^{**}p<0.01$, $^{***}p<0.001$.}
\centering
\scriptsize
\begin{tabular}[t]{llllllll}
\toprule
 & Estimate & SE & $t$ & LB$_{95}$ & UB$_{95}$ & $p$-value & Sig.\\
\midrule
\textsc{Neutral} × Post & $-$0.009 & 0.038 & $-$0.247 & $-$0.084 & 0.065 & 1.000 & \\
\textsc{Debunking} × Post & 0.037 & 0.039 & 0.946 & $-$0.040 & 0.113 & 0.784 & \\
\textsc{Interview} × Post & $-$0.008 & 0.036 & $-$0.229 & $-$0.080 & 0.063 & 1.000 & \\
\addlinespace
\textsc{Empathy} × Post & $-$0.021 & 0.037 & $-$0.551 & $-$0.094 & 0.053 & 1.000 & \\
\textsc{Roleplay} × Post & 0.036 & 0.036 & 1.010 & $-$0.035 & 0.108 & 0.784 & \\
\bottomrule
\end{tabular}
\end{table}

\begin{table}
\centering
\caption{\label{tab:s2-dv2-static}\textbf{Re-analysis  using \textsc{Static Information} as reference condition in Study~2 (refugee support).} Estimation result for linear mixed-effects model with a participant random intercept. Only the five Condition $\times$ Post interaction terms are reported; one-sided tests with Holm-correction. Significance levels: $^{\dagger}p<0.1$, $^{*}p<0.05$, $^{**}p<0.01$, $^{***}p<0.001$.}
\centering
\scriptsize
\begin{tabular}[t]{llllllll}
\toprule
 & Estimate & SE & $t$ & LB$_{95}$ & UB$_{95}$ & $p$-value & Sig.\\
\midrule
\textsc{Neutral} × Post & $-$0.042 & 0.029 & $-$1.453 & $-$0.098 & 0.015 & 0.926 & \\
\textsc{Debunking} × Post & 0.027 & 0.029 & 0.917 & $-$0.031 & 0.085 & 0.720 & \\
\textsc{Interview} × Post & 0.013 & 0.028 & 0.486 & $-$0.041 & 0.068 & 0.922 & \\
\addlinespace
\textsc{Empathy} × Post & 0.014 & 0.028 & 0.504 & $-$0.042 & 0.070 & 0.922 & \\
\textsc{Roleplay} × Post & 0.084 & 0.027 & 3.056 & 0.030 & 0.138 & 0.006 & **\\
\bottomrule
\end{tabular}
\end{table}

\begin{table}
\centering
\caption{\label{tab:s2-dv3-static}\textbf{Re-analysis  using \textsc{Static Information} as reference condition in Study~2 (policy endorsement).} Estimation result for linear mixed-effects model with a participant random intercept. Only the five Condition $\times$ Post interaction terms are reported; one-sided tests with Holm-correction. Significance levels: $^{\dagger}p<0.1$, $^{*}p<0.05$, $^{**}p<0.01$, $^{***}p<0.001$.}
\centering
\scriptsize
\begin{tabular}[t]{llllllll}
\toprule
 & Estimate & SE & $t$ & LB$_{95}$ & UB$_{95}$ & $p$-value & Sig.\\
\midrule
\textsc{Neutral} × Post & $-$0.021 & 0.025 & $-$0.853 & $-$0.070 & 0.028 & 1.000 & \\
\textsc{Debunking} × Post & 0.043 & 0.025 & 1.680 & $-$0.007 & 0.092 & 0.236 & \\
\textsc{Interview} × Post & $-$0.012 & 0.024 & $-$0.522 & -0.059 & 0.034 & 1.000 & \\
\addlinespace
\textsc{Empathy} × Post & 0.036 & 0.024 & 1.463 & $-$0.012 & 0.084 & 0.236 & \\
\textsc{Roleplay} × Post & 0.039 & 0.024 & 1.652 & $-$0.007 & 0.085 & 0.236 & \\
\bottomrule
\end{tabular}
\end{table}

\newpage

\subsection{Study~3}
\label{suppsec:roubstnessCheck_study3}

\begin{table}
\centering
\caption{\label{tab:s3-fu-dv1-Static information }\textbf{Re-analysis using \textsc{Static Information} as reference condition in Study 3 (perception).} Estimation result for OLS regression of perception at the two-week follow-up on condition. The five condition contrasts relative to \textsc{STATIC INFORMATION} are only reported; one-sided tests with Holm-corrected across the five contrasts. Significance levels: $^{\dagger}p<0.1$, $^{*}p<0.05$, $^{**}p<0.01$, $^{***}p<0.001$.}
\centering
\scriptsize
\begin{tabular}[t]{llllllll}
\toprule
Contrast & Estimate & SE & $t$ & LB$_{95}$ & UB$_{95}$ & $p$-value & Sig.\\
\midrule
\textsc{Neutral} -- \textsc{Static information} & 0.058 & 0.040 & 1.459 & $-$0.020 & 0.136 & 0.363 & \\
\textsc{Debunking} -- \textsc{Static information} & 0.057 & 0.040 & 1.420 & $-$0.022 & 0.136 & 0.363 & \\
\textsc{Interview} -- \textsc{Static information} & 0.034 & 0.040 & 0.844 & $-$0.045 & 0.114 & 0.399 & \\
\textsc{Empathy} -- \textsc{Static information} & 0.049 & 0.043 & 1.157 & $-$0.034 & 0.133 & 0.372 & \\
\textsc{Roleplay} -- \textsc{Static information} & 0.024 & 0.040 & 0.604 & $-$0.055 & 0.104 & 0.399 & \\
\bottomrule
\end{tabular}
\end{table}

\begin{table}
\centering
\caption{\label{tab:s3-fu-dv2-flyer}\textbf{Re-analysis  using \textsc{Static Information} as reference condition in Study 3 (refugee support).} Estimation result for OLS regression of refugee support at the two-week follow-up on condition. The five condition contrasts relative to \textsc{STATIC INFORMATION} are only reported; one-sided tests with Holm-corrected across the five contrasts. Significance levels: $^{\dagger}p<0.1$, $^{*}p<0.05$, $^{**}p<0.01$, $^{***}p<0.001$.}
\centering
\scriptsize
\begin{tabular}[t]{llllllll}
\toprule
Contrast & Estimate & SE & $t$ & LB$_{95}$ & UB$_{95}$ & $p$-value & Sig.\\
\midrule
\textsc{Neutral} -- \textsc{static information} & $-$0.043 & 0.032 & $-$1.326 & $-$0.107 & 0.021 & 1.000 & \\
\textsc{Debunking} -- \textsc{static information} & $-$0.021 & 0.033 & $-$0.634 & $-$0.085 & 0.043 & 1.000 & \\
\textsc{Interview} -- \textsc{static information} & $-$0.033 & 0.033 & $-$0.998 & $-$0.098 & 0.032 & 1.000 & \\
\textsc{Empathy} -- \textsc{static information} & $-$0.030 & 0.035 & $-$0.865 & $-$0.098 & 0.038 & 1.000 & \\
\textsc{Roleplay} -- \textsc{static information} & 0.016 & 0.033 & 0.496 & $-$0.048 & 0.081 & 1.000 & \\
\bottomrule
\end{tabular}
\end{table}

\begin{table}
\centering
\caption{\label{tab:s3-fu-dv3-flyer}\textbf{Re-analysis  using \textsc{Static Information} as reference condition in Study 3 (policy endorsement).} Estimation result for OLS regression of policy endorsement at the two-week follow-up on condition. The five condition contrasts relative to \textsc{STATIC INFORMATION} are only reported; one-sided tests with Holm-corrected across the five contrasts. Significance levels: $^{\dagger}p<0.1$, $^{*}p<0.05$, $^{**}p<0.01$, $^{***}p<0.001$.}
\centering
\scriptsize
\begin{tabular}[t]{llllllll}
\toprule
Contrast & Estimate & SE & $t$ & LB$_{95}$ & UB$_{95}$ & $p$-value & Sig.\\
\midrule
\textsc{Neutral} -- \textsc{static information} & $-$0.048 & 0.047 & $-$1.019 & $-$0.139 & 0.044 & 1.000 & \\
\textsc{Debunking} -- \textsc{static information} & $-$0.005 & 0.047 & $-$0.105 & $-$0.097 & 0.087 & 1.000 & \\
\textsc{Interview} -- \textsc{static information} & $-$0.006 & 0.047 & $-$0.126 & $-$0.099 & 0.087 & 1.000 & \\
\textsc{Empathy} -- \textsc{static information} & $-$0.025 & 0.050 & $-$0.508 & $-$0.123 & 0.073 & 1.000 & \\
\textsc{Roleplay} -- \textsc{static information} & 0.048 & 0.047 & 1.015 & $-$0.045 & 0.141 & 0.776 & \\
\bottomrule
\end{tabular}
\end{table}

\newpage

\section{Robustness to chat duration}
\label{suppsec:engagementControl}


The LLM interventions differed in their prescribed number of conversational rounds. To assess whether the estimated intervention effects were sensitive to differences in interaction length, we conducted a robustness check adjusting for chat duration. Chat duration was derived from server-side chatbot logs and measured as the elapsed time between the first and last recorded participant messages within each session, summed across sessions for each participant. Sessions with fewer than two recorded participant messages were removed. We report the results for Study~2 because we noticed a serve-side error in the other studies where, during periods of high several loads, not all conversations logs were stored correctly, which this affected only the conversation logs, not any of the variables.

To so so, we extended the main linear mixed-effects model (Supplement~\ref{suppsec:models}) by including chat duration as another control variable. Chat duration was standardized to a mean of zero and a standard deviation of one. The intervention effects thus represent differences in pre--post changes relative to \textsc{Neutral}, adjusted for chat duration. The results are in Supplementary~Table~\ref{tab:supp_engagement_condition_coefs_s2}. Overall, the estimation results after adjusting for chat duration support our main findings.

\begin{table}
\centering
\caption{\label{tab:supp_engagement_condition_coefs_s2}\textbf{Robustness check controlling for chat duration.} Estimation results for Condition-vs-\textsc{Neutral} in Study~2. The results are obtained separately with different models for the three dependent variables in the duration-adjusted models with one-sided tests with Holm-correction; 95\% confidence intervals. Significance levels: $^{\dagger}p<0.10$, $^{*}p<0.05$, $^{**}p<0.01$, $^{***}p<0.001$.}
\centering
\scriptsize
\begin{tabular}[t]{llllllll}
\toprule
Variable & Condition & Estimate & SE & $t$ & 95\% CI & $p$-value & Sig.\\
\midrule
\addlinespace[0.3em]
\emph{Model for perceptions} \\
\hspace{1em}Perceptions & \textsc{Debunking} & 0.047 & 0.040 & 1.174 & {}[$-$0.032, 0.125] & 0.434 &\\
\hspace{1em}Perceptions & \textsc{Interview} & 0.004 & 0.037 & 0.120 & {}[$-$0.069, 0.078] & 0.904 & \\
\hspace{1em}Perceptions & \textsc{Empathy} & $-$0.013 & 0.038 & $-$0.335 & {}[$-$0.089, 0.063] & 0.904 &\\
\hspace{1em}Perceptions & \textsc{Roleplay} & 0.047 & 0.038 & 1.238 & {}[$-$0.028, 0.121] & 0.434 & \\
\midrule
\emph{Model for refugee support} \\
\hspace{1em}Refugee support & \textsc{Debunking} & 0.069 & 0.031 & 2.264 & {}[0.009, 0.129] & 0.037 & *\\
\hspace{1em}Refugee support & \textsc{Interview} & 0.063 & 0.028 & 2.220 & {}[0.007, 0.119] & 0.037 & *\\
\hspace{1em}Refugee support & \textsc{Empathy} & 0.064 & 0.029 & 2.182 & {}[0.006, 0.122] & 0.037 & *\\
\hspace{1em}Refugee support & \textsc{Roleplay} & 0.126 & 0.029 & 4.352 & {}[0.069, 0.183] & $<$0.001 & ***\\
\midrule
\emph{Model for policy endorsement} \\
\hspace{1em}Policy endorsement & \textsc{Debunking} & 0.062 & 0.026 & 2.429 & {}[0.012, 0.112] & 0.024 & * \\
\hspace{1em}Policy endorsement & \textsc{Interview} & 0.010 & 0.024 & 0.422 & {}[$-$0.037, 0.057] & 0.337 & \\
\hspace{1em}Policy endorsement & \textsc{Empathy} & 0.063 & 0.025 & 2.579 & {}[0.015, 0.112] & 0.021 & *\\
\hspace{1em}Policy endorsement & \textsc{Roleplay} & 0.051 & 0.024 & 2.129 & {}[0.004, 0.099] & 0.034 & *\\
\bottomrule
\end{tabular}
\end{table}

\newpage

\section{Detailed results for Study 1}
\label{suppsec:study_1}

\subsection{Summary statistics}

\begin{table}[H]
\centering
\begin{table}
\centering
\caption{\textbf{Descriptive statistics for Study 1.} Reported: pre/post dependent variables. Note that all variables are normalized to the range 0--1.}
\label{tab:study1_descriptive_statistics_table}
\centering
\scriptsize
\begin{tabular}[t]{lSSSSSSSS}
\toprule
Variable & $n$ & {M} & {SD} & {Mdn} & {min} & {max} & {skew} & {kurtosis}\\
\midrule
Perceptions (pre) & 554 & 0.543 & 0.265 & 0.542 & 0 & 1.000 & -0.166 & -1.008\\
Perceptions (post) & 554 & 0.549 & 0.283 & 0.583 & 0 & 1.000 & -0.232 & -1.016\\
Refugee support (pre) & 554 & 0.468 & 0.217 & 0.470 & 0 & 0.963 & -0.037 & -0.573\\
Refugee support (post) & 554 & 0.498 & 0.249 & 0.506 & 0 & 1.000 & -0.095 & -0.763\\
Policy endorsement (pre) & 554 & 0.525 & 0.299 & 0.550 & 0 & 1.000 & -0.260 & -0.950\\
Policy endorsement (post) & 554 & 0.551 & 0.316 & 0.611 & 0 & 1.000 & -0.372 & -1.035\\
\bottomrule
\end{tabular}
\end{table}

\end{table}

\newpage

\subsection{Main effects}
\label{si:s1_confirmatory}

\subsubsection{Perceptions}

\begin{table}[H]
\centering
\begin{table}
\centering
\caption{\label{tab:s1-dv1-conf}\textbf{Perceptions: estimation results for the pre--post contrasts in Study~1} (i.e., interaction Condition $\times$ Post in the linear mixed-effects model; one-sided tests with Holm-correction; 95\% confidence intervals). 
Significance levels: $^{\dagger}p<0.1$, $^{*}p<0.05$, $^{**}p<0.01$, $^{***}p<0.001$.}
\centering
\scriptsize
\begin{tabular}[t]{llllllll}
\toprule
 & Estimate & SE & $t$ & LB$_{95}$ & UB$_{95}$ & $p$-value & Sig.\\
\midrule
\textsc{Debunking} × Post & $-$0.004 & 0.024 & $-$0.162 & $-$0.050 & 0.043 & 0.564 & \\
\textsc{Interview} × Post & 0.016 & 0.024 & 0.660 & $-$0.032 & 0.063 & 0.510 & \\
\textsc{Empathy} × Post & 0.041 & 0.023 & 1.781 & $-$0.004 & 0.087 & 0.151 & \\
\textsc{Roleplay} × Post & 0.039 & 0.024 & 1.640 & $-$0.008 & 0.087 & 0.153 & \\
\bottomrule
\end{tabular}
\end{table}

\end{table}

\subsubsection{Refugee support}

\begin{table}[H]
\centering
\begin{table}
\centering
\caption{\label{tab:s1-dv2-conf}\textbf{Refugee support: estimation results for the pre--post contrasts in Study~1} (i.e., interaction Condition $\times$ Post in the linear mixed-effects model; one-sided tests with Holm-correction; 95\% confidence intervals). 
Significance levels: $^{\dagger}p<0.1$, $^{*}p<0.05$, $^{**}p<0.01$, $^{***}p<0.001$.}
\centering
\scriptsize
\begin{tabular}[t]{llllllll}
\toprule
 & Estimate & SE & $t$ & LB$_{95}$ & UB$_{95}$ & $p$-value & Sig.\\
\midrule
\textsc{Debunking} × Post & 0.011 & 0.014 & 0.735 & -0.018 & 0.039 & 0.231 & \\
\textsc{Interview} × Post & 0.029 & 0.015 & 2.000 & 0.001 & 0.058 & 0.046 & *\\
\textsc{Empathy} × Post & 0.036 & 0.014 & 2.573 & 0.009 & 0.064 & 0.016 & *\\
\textsc{Roleplay} × Post & 0.097 & 0.015 & 6.651 & 0.068 & 0.126 & $<$0.001 & ***\\
\bottomrule
\end{tabular}
\end{table}

\end{table}

\subsubsection{Policy endorsement}

\begin{table}[H]
\centering
\begin{table}
\centering
\caption{\label{tab:s1-dv3-conf}\textbf{Policy endorsement: estimation results for the pre--post contrasts in Study~1} (i.e., interaction Condition $\times$ Post terms in the linear mixed-effects model; one-sided statistical test with Holm correction across the four contrasts; 95\% confidence intervals two-sided and unadjusted). $^{\dagger}p<0.1$, $^{*}p<0.05$, $^{**}p<0.01$, $^{***}p<0.001$.}
\centering
\scriptsize
\begin{tabular}[t]{llllllll}
\toprule
& Estimate & SE & $t$ & LB$_{95}$ & UB$_{95}$ & $p$-value & Sig.\\
\midrule
\textsc{Debunking} × Post & 0.013 & 0.012 & 1.100 & $-$0.010 & 0.036 & 0.272 & \\
\textsc{Interview} × Post & 0.010 & 0.012 & 0.812 & $-$0.014 & 0.033 & 0.272 & \\
\textsc{Empathy} × Post & 0.025 & 0.011 & 2.138 & 0.002 & 0.047 & 0.050 & *\\
\textsc{Roleplay} × Post & 0.057 & 0.012 & 4.824 & 0.034 & 0.081 & $<$0.001 & ***\\
\bottomrule
\end{tabular}
\end{table}

\end{table}

\subsection{Pairwise contrasts among interventions}
\label{suppsec:s1_pairwise}

\begin{table}[H]
\centering
\begin{table}
\centering
\caption{\textbf{Perceptions: differences between interventions in pre--post change in Study 1} (i.e., Condition $\times$ Time interaction; two-sided statistical test; Holm-correction). Significance thresholds: $^{\dagger}p<0.1$, $^{*} p < 0.05$, $^{**} p < 0.01$, $^{***} p < 0.001$.}
\scriptsize
\centering
\begin{tabular}[t]{lrrrrrrrl}
\toprule
Contrast & Estimate & SE & $df$ & $t$ & $p$-value & LB$_{95}$ & UB$_{95}$ & Sig.\\
\midrule
\textsc{Interview} -- \textsc{Debunking} & 0.020 & 0.025 & 453 & 0.795 & 1.000 & $-$0.029 & 0.069 & \\
\textsc{Empathy} -- \textsc{Debunking} & 0.045 & 0.024 & 453 & 1.886 & 0.359 & $-$0.002 & 0.092 & \\
\textsc{Empathy} -- \textsc{Interview} & 0.025 & 0.024 & 453 & 1.035 & 1.000 & $-$0.023 & 0.073 & \\
\textsc{Roleplay} -- \textsc{Debunking} & 0.043 & 0.025 & 453 & 1.747 & 0.406 & $-$0.005 & 0.092 & \\
\textsc{Roleplay} -- \textsc{Interview} & 0.023 & 0.025 & 453 & 0.928 & 1.000 & $-$0.026 & 0.073 & \\
\textsc{Roleplay} -- \textsc{Empathy} & $-$0.002 & 0.024 & 453 & $-$0.075 & 1.000 & $-$0.049 & 0.046 & \\
\bottomrule
\end{tabular}
\end{table}

\label{tab:study1_DV1_simple_effects_pairs}
\end{table}

\begin{table}[H]
\centering
\begin{table}
\centering
\caption{\textbf{Refugee support: differences between interventions in pre--post change in Study 1} (i.e., Condition $\times$ Time interaction; two-sided statistical test; Holm-correction). Significance thresholds: $^{\dagger}p<0.10$, $^{*} p < 0.05$, $^{**} p < 0.01$, $^{***} p < 0.001$.}
\centering
\scriptsize
\begin{tabular}[t]{lrrrrrrrl}
\toprule
Contrast & Estimate & SE & $df$ & $t$ & $p$-value & LB$_{95}$ & UB$_{95}$ & Sig.\\
\midrule
\textsc{Interview} -- \textsc{Debunking} & 0.019 & 0.015 & 453 & 1.244 & 0.429 & $-$0.011 & 0.049 & \\
\textsc{Empathy} -- \textsc{Debunking} & 0.026 & 0.015 & 453 & 1.764 & 0.235 & $-$0.003 & 0.054 & \\
\textsc{Empathy} -- \textsc{Interview} & 0.007 & 0.015 & 453 & 0.458 & 0.647 & $-$0.022 & 0.036 & \\
\textsc{Roleplay} -- \textsc{Debunking} & 0.087 & 0.015 & 453 & 5.755 & 0.000 & 0.057 & 0.116 & ***\\
\textsc{Roleplay} -- \textsc{Interview} & 0.068 & 0.015 & 453 & 4.415 & 0.000 & 0.038 & 0.098 & ***\\
\textsc{Roleplay} -- \textsc{Empathy} & 0.061 & 0.015 & 453 & 4.135 & 0.000 & 0.032 & 0.090 & ***\\
\bottomrule
\end{tabular}
\end{table}

\label{tab:study1_DV2_simple_effects_pairs}
\end{table}

\begin{table}[H]
\centering
\begin{table}
\centering
\caption{\textbf{Policy endorsement: differences between interventions in pre--post change in Study~1} (i.e., Condition $\times$ Time interaction; two-sided statistical test; Holm-correction). Significance thresholds: $^{\dagger}p<0.10$, $^{*} p < 0.05$, $^{**} p < 0.01$, $^{***} p < 0.001$.}
\scriptsize
\centering
\begin{tabular}[t]{lrrrrrrrl}
\toprule
Contrast & Estimate & SE & $df$ & t & $p$-value & LB$_{95}$ & UB$_{95}$ & Sig.\\
\midrule
\textsc{Interview} -- \textsc{Debunking} & $-$0.003 & 0.012 & 453 & $-$0.257 & 0.797 & $-$0.027 & 0.021 & \\
\textsc{Empathy} -- \textsc{Debunking} & 0.012 & 0.012 & 453 & 0.982 & 0.666 & $-$0.012 & 0.035 & \\
\textsc{Empathy} -- \textsc{Interview} & 0.015 & 0.012 & 453 & 1.223 & 0.666 & $-$0.009 & 0.039 & \\
\textsc{Roleplay} -- \textsc{Debunking} & 0.045 & 0.012 & 453 & 3.632 & 0.002 & 0.020 & 0.069 & **\\
\textsc{Roleplay} -- \textsc{Interview} & 0.048 & 0.013 & 453 & 3.813 & 0.001 & 0.023 & 0.072 & ***\\
\textsc{Roleplay} -- \textsc{Empathy} & 0.033 & 0.012 & 453 & 2.739 & 0.026 & 0.009 & 0.057 & *\\
\bottomrule
\end{tabular}
\end{table}

\label{tab:study1_DV3_simple_effects_pairs}
\end{table}

\clearpage

\section{Detailed results for Study 2}
\label{suppsec:study_2}
\subsection{Summary statistics}
\label{suppsec:study_2_summary_stat}

\begin{table}[H]
\centering
\begin{table}
\centering
\caption{\textbf{Descriptive statistics for pre/post dependent variables in Study~2.} Note that all variables are normalized to the range 0--1.}
\label{tab:study2_descriptive_statistics}
\centering
\scriptsize
\begin{tabular}[t]{lSSSSSSSS}
\toprule
Variable & $n$ & {M} & {SD} & {Mdn} & {min} & {max} & {skew} & {kurtosis}\\
\midrule
Perceptions (pre) & 250 & 0.688 & 0.195 & 0.708 & 0.125 & 1 & -0.445 & -0.482\\
Perceptions (post) & 250 & 0.712 & 0.215 & 0.750 & 0.000 & 1 & -0.594 & -0.239\\
Refugee support (pre) & 250 & 0.646 & 0.197 & 0.665 & 0.000 & 1 & -0.597 & 0.356\\
Refugee support (post) & 250 & 0.699 & 0.195 & 0.721 & 0.013 & 1 & -0.671 & 0.346\\
Policy endorsement (pre) & 250 & 0.747 & 0.211 & 0.786 & 0.040 & 1 & -0.839 & 0.487\\
Policy endorsement (post) & 250 & 0.795 & 0.201 & 0.840 & 0.000 & 1 & -1.139 & 1.182\\
\bottomrule
\end{tabular}
\end{table}

\end{table}

\subsection{Main effects}
\label{si:s2_confirmatory}

\subsubsection{Perceptions}
\begin{table}[H]
\centering
\begin{table}
\centering
\caption{\label{tab:s2-dv1-conf}\textbf{Perceptions: estimation results for the pre--post contrasts in Study~2} (i.e., interaction Condition $\times$ Post in the linear mixed-effects model; one-sided tests with Holm-correction; 95\% confidence intervals). 
Significance levels: $^{\dagger}p<0.1$, $^{*}p<0.05$, $^{**}p<0.01$, $^{***}p<0.001$.}
\centering
\scriptsize
\begin{tabular}[t]{llllllll}
\toprule
 & Estimate & SE & $t$ & LB$_{95}$ & UB$_{95}$ & $p$-value & Sig.\\
\midrule
\textsc{Debunking} × Post & 0.046 & 0.038 & 1.202 & $-$0.029 & 0.122 & 0.393 & \\
\textsc{Interview} × Post & 0.001 & 0.036 & 0.029 & $-$0.069 & 0.071 & 0.977 & \\
\textsc{Empathy} × Post & $-$0.011 & 0.037 & $-$0.307 & $-$0.084 & 0.061 & 0.977 & \\
\textsc{Roleplay} × Post & 0.046 & 0.035 & 1.295 & $-$0.024 & 0.116 & 0.393 & \\
\bottomrule
\end{tabular}
\end{table}

\label{tab:study2_interactions_DV1}
\end{table}

\subsubsection{Refugee support}

\begin{table}[H]
\centering
\begin{table}
\centering
\caption{\label{tab:s2-dv2-conf}\textbf{Refugee support: estimation results for the pre--post contrasts in Study~2} (i.e., interaction Condition $\times$ Post in the linear mixed-effects model; one-sided tests with Holm-correction; 95\% confidence intervals). 
Significance levels: $^{\dagger}p<0.10$, $^{*}p<0.05$, $^{**}p<0.01$, $^{***}p<0.001$.}
\centering
\scriptsize
\begin{tabular}[t]{llllllll}
\toprule
 & Estimate & SE & $t$ & LB$_{95}$ & UB$_{95}$ & $p$-value & Sig.\\
\midrule
\textsc{Debunking} × Post & 0.069 & 0.029 & 2.354 & 0.011 & 0.126 & 0.029 & *\\
\textsc{Interview} × Post & 0.055 & 0.027 & 2.034 & 0.002 & 0.109 & 0.043 & *\\
\textsc{Empathy} × Post & 0.056 & 0.028 & 1.997 & 0.001 & 0.112 & 0.043 & *\\
\textsc{Roleplay} × Post & 0.126 & 0.027 & 4.653 & 0.072 & 0.179 & $<$0.001 & ***\\
\bottomrule
\end{tabular}
\end{table}

\label{tab:study2_interactions_DV2}
\end{table}

\subsubsection{Policy endorsement}

\begin{table}[H]
\centering
\begin{table}
\centering
\caption{\label{tab:s2-dv3-conf}\textbf{Policy endorsement: estimation results for the pre--post contrasts in Study~2} (i.e., interaction Condition $\times$ Post in the linear mixed-effects model; one-sided tests with Holm-correction; 95\% confidence intervals). 
Significance levels: $^{\dagger}p<0.10$, $^{*}p<0.05$, $^{**}p<0.01$, $^{***}p<0.001$.}
\centering
\scriptsize
\begin{tabular}[t]{llllllll}
\toprule
 & Estimate & SE & $t$ & LB$_{95}$ & UB$_{95}$ & $p$-value & Sig.\\
\midrule
\textsc{Debunking} × Post & 0.064 & 0.025 & 2.585 & 0.015 & 0.112 & 0.018 & *\\
\textsc{Interview} × Post & 0.009 & 0.023 & 0.381 & $-$0.036 & 0.054 & 0.352 & \\
\textsc{Empathy} × Post & 0.057 & 0.024 & 2.402 & 0.010 & 0.104 & 0.018 & *\\
\textsc{Roleplay} × Post & 0.060 & 0.023 & 2.641 & 0.015 & 0.105 & 0.018 & *\\
\bottomrule
\end{tabular}
\end{table}

\label{tab:study2_interactions_DV3}
\end{table}

\subsection{Pairwise contrasts among interventions}
\label{suppsec:s2_pairwise}

\begin{table}[H]
\centering
\begin{table}
\centering
\caption{\textbf{Perceptions: differences between interventions in pre--post change in Study~2} (i.e., Condition $\times$ Time interaction; two-sided statistical test; Holm-correction). Significance thresholds: $^{\dagger}p<0.1$, $^{*} p < 0.05$, $^{**} p < 0.01$, $^{***} p < 0.001$.}
\centering
\scriptsize
\begin{tabular}[t]{lrrrrrrrl}
\toprule
Contrast & Estimate & SE & $df$ & $t$ & $p$-value & LB$_{95}$ & UB$_{95}$ & Sig.\\
\midrule
\textsc{Interview} -- \textsc{Debunking} & $-$0.045 & 0.037 & 212 & $-$1.230 & 0.731 & $-$0.117 & 0.027 & \\
\textsc{Empathy} -- \textsc{Debunking} & $-$0.057 & 0.038 & 212 & $-$1.514 & 0.657 & $-$0.132 & 0.017 & \\
\textsc{Empathy} -- \textsc{Interview} & $-$0.012 & 0.035 & 212 & $-$0.352 & 1.000 & $-$0.081 & 0.057 & \\
\textsc{Roleplay} -- \textsc{Debunking} & 0.000 & 0.036 & 212 & $-$0.006 & 1.000 & $-$0.072 & 0.072 & \\
\textsc{Roleplay} -- \textsc{Interview} & 0.045 & 0.034 & 212 & 1.336 & 0.731 & $-$0.021 & 0.111 & \\
\textsc{Roleplay} -- \textsc{Empathy} & 0.057 & 0.035 & 212 & 1.637 & 0.619 & $-$0.012 & 0.126 & \\
\bottomrule
\end{tabular}
\end{table}

\label{tab:study2_simple_effects_DV1}
\end{table}

\begin{table}[H]
\centering
\begin{table}
\centering
\caption{\textbf{Refugee support: differences between interventions in pre--post change in Study~2} (i.e., Condition $\times$ Time interaction; two-sided statistical test; Holm-correction). Significance thresholds: $^{\dagger}p<0.1$, $^{*} p < 0.05$, $^{**} p < 0.01$, $^{***} p < 0.001$.}
\scriptsize
\centering
\begin{tabular}[t]{lrrrrrrrl}
\toprule
Contrast & Estimate & SE & $df$ & $t$ & $p$-value & LB$_{95}$ & UB$_{95}$ & Sig.\\
\midrule
\textsc{Interview} -- \textsc{Debunking} & $-$0.014 & 0.028 & 212 & $-$0.489 & 1.000 & $-$0.069 & 0.041 & \\
\textsc{Empathy} -- \textsc{Debunking} & $-$0.013 & 0.029 & 212 & $-$0.439 & 1.000 & $-$0.070 & 0.044 & \\
\textsc{Empathy} -- \textsc{Interview} & 0.001 & 0.027 & 212 & 0.036 & 1.000 & $-$0.052 & 0.054 & \\
\textsc{Roleplay} -- \textsc{Debunking} & 0.057 & 0.028 & 212 & 2.041 & 0.170 & 0.002 & 0.112 & \\
\textsc{Roleplay} -- \textsc{Interview} & 0.070 & 0.026 & 212 & 2.753 & 0.038 & 0.020 & 0.121 & *\\
\textsc{Roleplay} -- \textsc{Empathy} & 0.069 & 0.027 & 212 & 2.608 & 0.049 & 0.017 & 0.122 & *\\
\bottomrule
\end{tabular}
\end{table}

\label{tab:study2_simple_effects_DV2}
\end{table}

\begin{table}[H]
\centering
\begin{table}
\centering
\caption{\textbf{Policy endorsement: differences between interventions in pre--post change in Study~2} (i.e., Condition $\times$ Time interaction; two-sided statistical test; Holm-correction). Significance thresholds: $^{\dagger}p<0.1$, $^{*} p < 0.05$, $^{**} p < 0.01$, $^{***} p < 0.001$.}
\centering
\scriptsize
\begin{tabular}[t]{lrrrrrrrl}
\toprule
Contrast & Estimate & SE & $df$ & $t$ & $p$-value & LB$_{95}$ & UB$_{95}$ & Sig.\\
\midrule
\textsc{Interview} -- \textsc{Debunking} & $-$0.055 & 0.024 & 212 & $-$2.335 & 0.108 & $-$0.101 & $-$0.009 & \\
\textsc{Empathy} -- \textsc{Debunking} & $-$0.007 & 0.024 & 212 & $-$0.278 & 1.000 & $-$0.055 & 0.041 & \\
\textsc{Empathy} -- \textsc{Interview} & 0.048 & 0.023 & 212 & 2.137 & 0.135 & 0.004 & 0.093 & \\
\textsc{Roleplay} -- \textsc{Debunking} & $-$0.004 & 0.023 & 212 & $-$0.154 & 1.000 & $-$0.050 & 0.043 & \\
\textsc{Roleplay} -- \textsc{Interview} & 0.051 & 0.022 & 212 & 2.383 & 0.108 & 0.009 & 0.094 & \\
\textsc{Roleplay} -- \textsc{Empathy} & 0.003 & 0.022 & 212 & 0.141 & 1.000 & $-$0.041 & 0.047 & \\
\bottomrule
\end{tabular}
\end{table}

\label{tab:study2_simple_effects_DV3}
\end{table}

\clearpage

\section{Detailed results for Study 3}
\label{suppsec:study_3}
\subsection{Summary statistics}
\label{suppsec:s3_descriptive}

\begin{table}
\centering
\caption{\textbf{Descriptive statistics for follow-up dependent variables in Study~3.} Note: dependent variables are normalized to range 0--1.}
\label{tab:descriptives_study3_follow}
\centering
\scriptsize
\begin{tabular}[t]{lrrrrrrrr}
\toprule
Variable & $n$ & {M} & {SD} & {Mdn} & {min} & {max} & {skew} & {kurtosis}\\
\midrule
Perceptions (follow-up) & 521 & 0.542 & 0.260 & 0.542 & 0 & 1.00 & -0.060 & -0.948\\
Refugee support (follow-up) & 521 & 0.520 & 0.213 & 0.525 & 0 & 0.98 & -0.178 & -0.603\\
Policy endorsement (follow-up) & 521 & 0.555 & 0.306 & 0.588 & 0 & 1.00 & -0.259 & -1.014\\
\bottomrule
\end{tabular}
\end{table}

\subsection{Main effects}
\label{suppsec:s3_confirmatory}

\subsubsection{Immediate post-intervention}
\label{suppsec:s3_confirmatory_post}

\begin{table}
\centering
\caption{\label{tab:s3-post-dv1}\textbf{Perceptions: estimation results for treatment-vs-\textsc{Neutral} contrasts at immediate post-intervention in Study~3.} OLS regression of the outcome on condition, \textsc{Neutral} as reference; one-sided tests with Holm-correction across the four contrasts. Significance thresholds: $^{\dagger}p<0.1$, $^{*}p<0.05$, $^{**}p<0.01$, $^{***}p<0.001$.}
\centering
\scriptsize
\begin{tabular}[t]{llllllll}
\toprule
Contrast & Estimate & SE & $t$ & LB$_{95}$ & UB$_{95}$ & $p$-value & Sig.\\
\midrule
\textsc{Debunking} -- \textsc{Neutral} & 0.001 & 0.037 & 0.015 & $-$0.072 & 0.073 & 1.000 & \\
\textsc{Interview} -- \textsc{Neutral} & $-$0.021 & 0.038 & $-$0.548 & $-$0.095 & 0.053 & 1.000 & \\
\textsc{Empathy} -- \textsc{Neutral} & $-$0.002 & 0.040 & $-$0.047 & $-$0.080 & 0.077 & 1.000 & \\
\textsc{Roleplay} -- \textsc{Neutral} & $-$0.040 & 0.038 & $-$1.054 & $-$0.114 & 0.034 & 1.000 & \\
\bottomrule
\end{tabular}
\end{table}

\begin{table}
\centering
\caption{\label{tab:s3-post-dv2}\textbf{Refugee support: estimation results for treatment-vs-\textsc{Neutral} contrasts at immediate post-intervention in Study~3.} OLS regression of the outcome on condition, \textsc{Neutral} as reference; one-sided tests with Holm-correction across the four contrasts. Significance thresholds: $^{\dagger}p<0.1$, $^{*}p<0.05$, $^{**}p<0.01$, $^{***}p<0.001$.}
\centering
\scriptsize
\begin{tabular}[t]{llllllll}
\toprule
Contrast & Estimate & SE & $t$ & LB$_{95}$ & UB$_{95}$ & $p$-value & Sig.\\
\midrule
\textsc{Debunking} -- \textsc{Neutral} & 0.024 & 0.030 & 0.787 & $-$0.035 & 0.083 & 0.647 & \\
\textsc{Interview} -- \textsc{Neutral} & 0.014 & 0.031 & 0.452 & $-$0.046 & 0.074 & 0.652 & \\
\textsc{Empathy} -- \textsc{Neutral} & $-$0.022 & 0.032 & $-$0.672 & $-$0.085 & 0.042 & 0.749 & \\
\textsc{Roleplay} -- \textsc{Neutral} & 0.077 & 0.031 & 2.532 & 0.017 & 0.138 & 0.023 & *\\
\bottomrule
\end{tabular}
\end{table}

\begin{table}
\centering
\caption{\label{tab:s3-post-dv3}\textbf{Policy endorsement: estimation results for treatment-vs-\textsc{Neutral} contrasts at immediate post-intervention in Study~3.} OLS regression of the outcome on condition, \textsc{Neutral} as reference; one-sided tests and Holm-correction across the four contrasts. Significance thresholds: $^{\dagger}p<0.1$, $^{*}p<0.05$, $^{**}p<0.01$, $^{***}p<0.001$.}
\centering
\scriptsize
\begin{tabular}[t]{llllllll}
\toprule
Contrast & Estimate & SE & $t$ & LB$_{95}$ & UB$_{95}$ & $p$-value & Sig.\\
\midrule
\textsc{Debunking} -- \textsc{Neutral} & 0.051 & 0.044 & 1.162 & $-$0.035 & 0.137 & 0.369 & \\
\textsc{Interview} -- \textsc{Neutral} & 0.037 & 0.045 & 0.835 & $-$0.050 & 0.125 & 0.404 & \\
\textsc{Empathy} -- \textsc{Neutral} & 0.036 & 0.047 & 0.757 & $-$0.057 & 0.129 & 0.404 & \\
\textsc{Roleplay} -- \textsc{Neutral} & 0.076 & 0.045 & 1.706 & $-$0.012 & 0.164 & 0.177 & \\
\bottomrule
\end{tabular}
\end{table}

\subsubsection{Two-week follow-up}
\label{suppsec:s3_confirmatory_fu}

\begin{table}
\centering
\caption{\label{tab:s3-fu-dv1}\textbf{Perceptions: estimation results for treatment-vs-\textsc{Neutral} contrasts at the two-week follow-up in Study~3}. OLS regression of the outcome on condition, \textsc{Neutral} as reference; $n = 444$; one-sided tests with Holm-correction across the four contrasts. Significance thresholds: $^{\dagger}p<0.1$, $^{*}p<0.05$, $^{**}p<0.01$, $^{***}p<0.001$.}
\centering
\scriptsize
\begin{tabular}[t]{llllllll}
\toprule
Contrast & Estimate & SE & $t$ & LB$_{95}$ & UB$_{95}$ & $p$-value & Sig.\\
\midrule
\textsc{Debunking} -- \textsc{Neutral} & $-$0.001 & 0.038 & $-$0.030 & $-$0.076 & 0.074 & 1.000 & \\
\textsc{Interview} -- \textsc{Neutral} & $-$0.024 & 0.039 & -0.620 & $-$0.100 & 0.052 & 1.000 & \\
\textsc{Empathy} -- \textsc{Neutral} & $-$0.009 & 0.041 & $-$0.215 & $-$0.089 & 0.071 & 1.000 & \\
\textsc{Roleplay} -- \textsc{Neutral} & $-$0.034 & 0.039 & $-$0.872 & $-$0.109 & 0.042 & 1.000 & \\
\bottomrule
\end{tabular}
\end{table}

\begin{table}
\centering
\caption{\label{tab:s3-fu-dv2}\textbf{Refugee support: estimation results for treatment-vs-\textsc{Neutral} contrasts at the two-week follow-up in Study~3.} OLS regression of the outcome on condition, \textsc{Neutral} as reference; $n = 444$; one-sided tests with Holm-corrected across the four contrasts. Significance thresholds: $^{\dagger}p<0.1$, $^{*}p<0.05$, $^{**}p<0.01$, $^{***}p<0.001$.}
\centering
\scriptsize
\begin{tabular}[t]{llllllll}
\toprule
Contrast & Estimate & SE & $t$ & LB$_{95}$ & UB$_{95}$ & $p$-value & Sig.\\
\midrule
\textsc{Debunking} -- \textsc{Neutral} & 0.022 & 0.031 & 0.719 & $-$0.039 & 0.083 & 0.708 & \\
\textsc{Interview} -- \textsc{Neutral} & 0.010 & 0.031 & 0.322 & $-$0.052 & 0.072 & 0.708 & \\
\textsc{Empathy} -- \textsc{Neutral} & 0.013 & 0.033 & 0.391 & $-$0.052 & 0.078 & 0.708 & \\
\textsc{Roleplay} -- \textsc{Neutral} & 0.059 & 0.031 & 1.895 & $-$0.002 & 0.121 & 0.118 & \\
\bottomrule
\end{tabular}
\end{table}

\begin{table}
\centering
\caption{\label{tab:s3-fu-dv3}\textbf{Policy endorsement: estimation results for treatment-vs-\textsc{Neutral} contrasts at the two-week follow-up in Study~3}. OLS regression of the outcome on condition, \textsc{Neutral} as reference; $n = 444$; one-sided tests and Holm-correction across the four contrasts. Significance thresholds: $^{\dagger}p<0.1$, $^{*}p<0.05$, $^{**}p<0.01$, $^{***}p<0.001$.}
\centering
\scriptsize
\begin{tabular}[t]{llllllll}
\toprule
Contrast & Estimate & SE & $t$ & LB$_{95}$ & UB$_{95}$ & $p$-value & Sig.\\
\midrule
\textsc{Debunking} -- \textsc{Neutral} & 0.043 & 0.045 & 0.953 & $-$0.045 & 0.130 & 0.511 & \\
\textsc{Interview} -- \textsc{Neutral} & 0.042 & 0.045 & 0.920 & $-$0.047 & 0.130 & 0.511 & \\
\textsc{Empathy} -- \textsc{Neutral} & 0.022 & 0.048 & 0.464 & $-$0.072 & 0.116 & 0.511 & \\
\textsc{Roleplay} -- \textsc{Neutral} & 0.096 & 0.045 & 2.118 & 0.007 & 0.185 & 0.070 & $\dagger$\\
\bottomrule
\end{tabular}
\end{table}

\subsection{Pairwise contrasts among interventions}
\label{suppsec:s3_pairwise}

\begin{table}
\centering
\caption{\label{tab:s3-post-dv1-pairwise}\textbf{Perceptions (immediate post-intervention): pairwise contrasts among interventions.} Two-sided tests with Holm-correction across the six intervention pairs. Significance thresholds: $^{\dagger}p<0.1$, $^{*}p<0.05$, $^{**}p<0.01$, $^{***}p<0.001$.}
\centering
\scriptsize
\begin{tabular}[t]{lrrrrrrll}
\toprule
Contrast & Estimate & SE & $df$ & $t$ & LB$_{95}$ & UB$_{95}$ & $p$-value & Sig.\\
\midrule
\textsc{Debunking} - \textsc{Empathy} & 0.002 & 0.040 & 460 & 0.061 & -0.077 & 0.081 & 1.000 & \\
\textsc{Debunking} - \textsc{Interview} & 0.021 & 0.038 & 460 & 0.560 & -0.054 & 0.096 & 1.000 & \\
\textsc{Debunking} - \textsc{Roleplay} & 0.040 & 0.038 & 460 & 1.064 & -0.035 & 0.115 & 1.000 & \\
\textsc{Empathy} - \textsc{Interview} & 0.019 & 0.041 & 460 & 0.462 & -0.062 & 0.100 & 1.000 & \\
\textsc{Empathy} - \textsc{Roleplay} & 0.038 & 0.041 & 460 & 0.931 & -0.043 & 0.119 & 1.000 & \\
\textsc{Interview} - \textsc{Roleplay} & 0.019 & 0.039 & 460 & 0.497 & -0.058 & 0.096 & 1.000 & \\
\bottomrule
\end{tabular}
\end{table}

\begin{table}
\centering
\caption{\label{tab:s3-post-dv2-pairwise}\textbf{Refugee support (immediate post-intervention): pairwise contrasts among interventions.} Two-sided tests with Holm-correction across the six intervention pairs. Significance thresholds: $^{\dagger}p<0.10$, $^{*}p<0.05$, $^{**}p<0.01$, $^{***}p<0.001$.}
\centering
\scriptsize
\begin{tabular}[t]{lrrrrrrll}
\toprule
Contrast & Estimate & SE & $df$ & $t$ & LB$_{95}$ & UB$_{95}$ & $p$-value & Sig.\\
\midrule
\textsc{Debunking} -- \textsc{Empathy} & 0.045 & 0.032 & 460 & 1.397 & -0.018 & 0.108 & 0.489 & \\
\textsc{Debunking} -- \textsc{Interview} & 0.010 & 0.031 & 460 & 0.321 & $-$0.051 & 0.071 & 0.748 & \\
\textsc{Debunking} -- \textsc{Roleplay} & $-$0.054 & 0.031 & 460 & $-$1.751 & $-$0.115 & 0.007 & 0.323 & \\
\textsc{Empathy} -- \textsc{Interview} & $-$0.036 & 0.033 & 460 & $-$1.079 & $-$0.101 & 0.029 & 0.563 & \\
\textsc{Empathy} -- \textsc{Roleplay} & $-$0.099 & 0.033 & 460 & $-$3.006 & $-$0.164 & -0.034 & 0.017 & *\\
\textsc{Interview} -- \textsc{Roleplay} & $-$0.064 & 0.031 & 460 & $-$2.040 & $-$0.125 & -0.003 & 0.210 & \\
\bottomrule
\end{tabular}
\end{table}

\begin{table}
\centering
\caption{\label{tab:s3-post-dv3-pairwise}\textbf{Policy endorsement (immediate post-intervention): pairwise contrasts among interventions.} Two-sided tests with Holm-correction across the six intervention pairs. Significance thresholds: $^{\dagger}p<0.10$, $^{*}p<0.05$, $^{**}p<0.01$, $^{***}p<0.001$.}
\centering
\scriptsize
\begin{tabular}[t]{lrrrrrrll}
\toprule
Contrast & Estimate & SE & $df$ & $t$ & LB$_{95}$ & UB$_{95}$ & $p$-value & Sig.\\
\midrule
\textsc{Debunking} -- \textsc{Empathy} & 0.015 & 0.048 & 460 & 0.321 & $-$0.079 & 0.109 & 1.000 & \\
\textsc{Debunking} -- \textsc{Interview} & 0.014 & 0.045 & 460 & 0.308 & $-$0.074 & 0.102 & 1.000 & \\
\textsc{Debunking} -- \textsc{Roleplay} & $-$0.025 & 0.045 & 460 & $-$0.563 & $-$0.113 & 0.063 & 1.000 & \\
\textsc{Empathy} -- \textsc{Interview} & $-$0.001 & 0.048 & 460 & $-$0.030 & $-$0.095 & 0.093 & 1.000 & \\
\textsc{Empathy} -- \textsc{Roleplay} & $-$0.041 & 0.048 & 460 & $-$0.840 & $-$0.135 & 0.053 & 1.000 & \\
\textsc{Interview} -- \textsc{Roleplay} & $-$0.039 & 0.046 & 460 & $-$0.856 & $-$0.129 & 0.051 & 1.000 & \\
\bottomrule
\end{tabular}
\end{table}

\begin{table}
\centering
\caption{\label{tab:s3-fu-dv1-pairwise}\textbf{Perceptions (follow-up): pairwise contrasts among interventions.} Two-sided tests with Holm-correction across the six intervention pairs. Significance thresholds: $^{\dagger}p<0.1$, $^{*}p<0.05$, $^{**}p<0.01$, $^{***}p<0.001$.}
\centering
\scriptsize
\begin{tabular}[t]{lrrrrrrll}
\toprule
Contrast & Estimate & SE & $df$ & $t$ & LB$_{95}$ & UB$_{95}$ & $p$-value & Sig.\\
\midrule
\textsc{Debunking} -- \textsc{Empathy} & 0.008 & 0.041 & 439 & 0.186 & $-$0.073 & 0.089 & 1.000 & \\
\textsc{Debunking} -- \textsc{Interview} & 0.023 & 0.039 & 439 & 0.586 & $-$0.054 & 0.100 & 1.000 & \\
\textsc{Debunking} -- \textsc{Roleplay} & 0.032 & 0.039 & 439 & 0.836 & $-$0.045 & 0.109 & 1.000 & \\
\textsc{Empathy} -- \textsc{Interview} & 0.015 & 0.041 & 439 & 0.364 & $-$0.066 & 0.096 & 1.000 & \\
\textsc{Empathy} -- \textsc{Roleplay} & 0.025 & 0.041 & 439 & 0.599 & $-$0.056 & 0.106 & 1.000 & \\
\textsc{Interview} -- \textsc{Roleplay} & 0.010 & 0.039 & 439 & 0.248 & $-$0.067 & 0.087 & 1.000 & \\
\bottomrule
\end{tabular}
\end{table}

\begin{table}
\centering
\caption{\label{tab:s3-fu-dv2-pairwise}\textbf{Refugee support (follow-up): pairwise contrasts among interventions.} Two-sided tests with Holm-correction across the six intervention pairs. Significance thresholds: $^{\dagger}p<0.1$, $^{*}p<0.05$, $^{**}p<0.01$, $^{***}p<0.001$.}
\centering
\scriptsize
\begin{tabular}[t]{lrrrrrrll}
\toprule
Contrast & Estimate & SE & $df$ & $t$ & LB$_{95}$ & UB$_{95}$ & $p$-value & Sig.\\
\midrule
\textsc{Debunking} -- \textsc{Empathy} & 0.009 & 0.033 & 439 & 0.279 & $-$0.056 & 0.074 & 1.000 & \\
\textsc{Debunking} -- \textsc{Interview} & 0.012 & 0.032 & 439 & 0.387 & $-$0.051 & 0.075 & 1.000 & \\
\textsc{Debunking} -- \textsc{Roleplay} & $-$0.037 & 0.032 & 439 & $-$1.174 & $-$0.100 & 0.026 & 0.964 & \\
\textsc{Empathy} -- \textsc{Interview} & 0.003 & 0.034 & 439 & 0.086 & $-$0.064 & 0.070 & 1.000 & \\
\textsc{Empathy} -- \textsc{Roleplay} & $-$0.046 & 0.034 & 439 & $-$1.376 & $-$0.113 & 0.021 & 0.848 & \\
\textsc{Interview} -- \textsc{Roleplay} & -0.049 & 0.032 & 439 & $-$1.544 & $-$0.112 & 0.014 & 0.739 & \\
\bottomrule
\end{tabular}
\end{table}

\begin{table}
\centering
\caption{\label{tab:s3-fu-dv3-pairwise}\textbf{Policy endorsement (follow-up): pairwise contrasts among interventions.} Two-sided tests with Holm-correction across the six intervention pairs. Significance thresholds: $^{\dagger}p<0.1$, $^{*}p<0.05$, $^{**}p<0.01$, $^{***}p<0.001$.}
\centering
\scriptsize
\begin{tabular}[t]{lrrrrrrll}
\toprule
Contrast & Estimate & SE & $df$ & $t$ & LB$_{95}$ & UB$_{95}$ & $p$-value & Sig.\\
\midrule
\textsc{Debunking} -- \textsc{Empathy} & 0.020 & 0.048 & 439 & 0.423 & $-$0.074 & 0.114 & 1.000 & \\
\textsc{Debunking} -- \textsc{Interview} & 0.001 & 0.046 & 439 & 0.023 & $-$0.089 & 0.091 & 1.000 & \\
\textsc{Debunking} -- \textsc{Roleplay} & $-$0.053 & 0.046 & 439 & $-$1.166 & $-$0.143 & 0.037 & 1.000 & \\
\textsc{Empathy} -- \textsc{Interview} & $-$0.019 & 0.049 & 439 & $-$0.398 & $-$0.115 & 0.077 & 1.000 & \\
\textsc{Empathy} -- \textsc{Roleplay} & $-$0.073 & 0.049 & 439 & $-$1.511 & $-$0.169 & 0.023 & 0.789 & \\
\textsc{Interview} -- \textsc{Roleplay} & -0.054 & 0.046 & 439 & $-$1.176 & $-$0.144 & 0.036 & 1.000 & \\
\bottomrule
\end{tabular}
\end{table}

\subsection{Durability: change from immediate post-intervention to follow-up}
\label{suppsec:s3-durability}

In the main text, the Study~3 effects are estimated with separate OLS regressions at each time point, thus testing whether each intervention exceeds \textsc{Neutral} immediately after the intervention or at the two-week follow-up. Here, we instead test the durability of these effects, i.e., whether the decay from the immediate post-intervention measurement to the follow-up is smaller in each intervention than in \textsc{Neutral}. For this, we fitted the linear mixed-effects model with a Condition~$\times$~Follow-Up interaction and a participant random intercept using the observation for participants with both measurements ($n = 444$; \textsc{Static Information} excluded). We tested whether each intervention showed a smaller decay than \textsc{Neutral} from the immediate post-intervention measurement to the follow-up, using one-sided tests with Holm correction across the four interventions within each outcome. The results are in Supplementary~Table~\ref{tab:s3-dur-change}.

\begin{table}
\centering
\caption{\label{tab:s3-dur-change}\textbf{Study 3: durability: change from immediate post-intervention to follow-up.} Change in the Condition-vs-\textsc{Neutral} difference from the immediate post-intervention measurement to the two-week follow-up (Condition $\times$ Follow-Up coefficients; $n = 444$; \textsc{Static Information} excluded). Positive values indicate a smaller decay than in \textsc{Neutral}. One-sided tests (smaller decay than \textsc{Neutral}) with Holm-correction across the four contrasts within outcome. Significance thresholds: $^{\dagger}p<0.10$, $^{*}p<0.05$, $^{**}p<0.01$, $^{***}p<0.001$.}
\centering
\scriptsize
\begin{tabular}[t]{llrrrrrrll}
\toprule
\multicolumn{2}{c}{ } & \multicolumn{8}{c}{Change (Follow-up $-$ Post)} \\
\cmidrule(l{3pt}r{3pt}){3-10}
Variable & Contrast & $\delta_g$ & SE & $df$ & $t$ & LB$_{95}$ & UB$_{95}$ & $p$-value & Sig.\\
\midrule
\emph{Model for policy perceptions} \\
Perceptions & \textsc{Debunking} $-$ \textsc{Neutral} & $-$0.012 & 0.036 & 439 & $-$0.317 & $-$0.083 & 0.060 & 1.000 & \\
Perceptions & \textsc{Empathy} $-$ \textsc{Neutral} & $-$0.011 & 0.039 & 439 & $-$0.276 & $-$0.087 & 0.066 & 1.000 & \\
Perceptions & \textsc{Interview} $-$ \textsc{Neutral} & $-$0.002 & 0.037 & 439 & $-$0.062 & $-$0.074 & 0.070 & 1.000 & \\
Perceptions & \textsc{Roleplay} $-$ \textsc{Neutral} & $-$0.001 & 0.037 & 439 & $-$0.024 & $-$0.073 & 0.071 & 1.000 & \\
\midrule
\emph{Model for refugee support} \\
Refugee support & \textsc{Debunking} $-$ \textsc{Neutral} & $-$0.003 & 0.018 & 439 & $-$0.159 & $-$0.037 & 0.032 & 1.000 & \\
Refugee support & \textsc{Empathy} $-$ \textsc{Neutral} & 0.022 & 0.019 & 439 & 1.189 & $-$0.015 & 0.059 & 0.470 & \\
Refugee support & \textsc{Interview} $-$ \textsc{Neutral} & $-$0.012 & 0.018 & 439 & $-$0.693 & $-$0.047 & 0.023 & 1.000 & \\
Refugee support & \textsc{Roleplay} $-$ \textsc{Neutral} & $-$0.028 & 0.018 & 439 & $-$1.601 & $-$0.063 & 0.006 & 1.000 & \\
\midrule
\emph{Model for policy endorsement} \\
Policy endorsement & \textsc{Debunking} $-$ \textsc{Neutral} & $-$0.009 & 0.018 & 439 & $-$0.464 & -0.045 & 0.028 & 1.000 & \\
Policy endorsement & \textsc{Empathy} $-$ \textsc{Neutral} & $-$0.029 & 0.020 & 439 & $-$1.463 & $-$0.068 & 0.010 & 1.000 & \\
Policy endorsement & \textsc{Interview} $-$ \textsc{Neutral} & -0.007 & 0.019 & 439 & $-$0.392 & $-$0.044 & 0.029 & 1.000 & \\
Policy endorsement & \textsc{Roleplay} $-$ \textsc{Neutral} & 0.006 & 0.019 & 439 & 0.327 & $-$0.031 & 0.043 & 1.000 & \\
\bottomrule
\end{tabular}
\end{table}

\clearpage

\subsection{Sensitivity across Holm-correction}
\label{suppsec:study3_alpha010}

\begin{table}[H]
\centering
\begin{table}
\centering
\caption{\label{tab:supp_alpha10_dv1}\textbf{Perceptions (follow-up): multiple-testing correction.}
Contrasts Condition-vs-\textsc{Neutral} with unadjusted and Holm-adjusted $p$-values at the $\alpha = .10$ threshold.}
\centering
\scriptsize
\begin{tabular}[t]{lrrrrrllll}
\toprule
Contrast & Estimate & SE & $df$ & LB$_{90}$ & $t$ & $p$-value & $p_{\mathrm{Holm}}$-value & Sig.\ at .10 (unadj.) & Sig.\ at .10 (Holm)\\
\midrule
\textsc{Debunking} -- \textsc{Neutral} & $-$0.001 & 0.038 & 439 & $-$0.050 & $-$0.030 & 0.512 & 1.000 & no & no\\
\textsc{Empathy} -- \textsc{Neutral} & $-$0.009 & 0.041 & 439 & $-$0.061 & $-$0.215 & 0.585 & 1.000 & no & no\\
\textsc{Interview} -- \textsc{Neutral} & $-$0.024 & 0.039 & 439 & $-$0.073 & $-$0.620 & 0.732 & 1.000 & no & no\\
\textsc{Roleplay} -- \textsc{Neutral} & $-$0.034 & 0.039 & 439 & $-$0.083 & $-$0.872 & 0.808 & 1.000 & no & no\\
\bottomrule
\end{tabular}
\end{table}

\end{table}

\begin{table}[H]
\centering
\begin{table}
\centering
\caption{\label{tab:supp_alpha10_dv2}\textbf{Refugee support (follow-up): multiple-testing correction..}
Contrasts Condition-vs-\textsc{Neutral} with unadjusted and Holm-adjusted $p$-values at the $\alpha = .10$ threshold.}
\centering
\scriptsize
\begin{tabular}[t]{lrrrrrllll}
\toprule
Contrast & Estimate & SE & $df$ & LB$_{90}$ & $t$ & $p$-value & $p_{\mathrm{Holm}}$-value & Sig.\ at .10 (unadj.) & Sig.\ at .10 (Holm)\\
\midrule
\textsc{Debunking} -- \textsc{Neutral} & 0.022 & 0.031 & 439 & $-$0.017 & 0.719 & 0.236 & 0.708 & no & no\\
\textsc{Empathy} -- \textsc{Neutral} & 0.013 & 0.033 & 439 & $-$0.030 & 0.391 & 0.348 & 0.708 & no & no\\
\textsc{Interview} -- \textsc{Neutral} & 0.010 & 0.031 & 439 & $-$0.030 & 0.322 & 0.374 & 0.708 & no & no\\
\textsc{Roleplay} -- \textsc{Neutral} & 0.059 & 0.031 & 439 & 0.019 & 1.895 & 0.029 & 0.118 & yes & no\\
\bottomrule
\end{tabular}
\end{table}

\end{table}

\begin{table}[H]
\centering
\begin{table}
\centering
\caption{\label{tab:supp_alpha10_dv3}\textbf{Policy endorsement (follow-up): multiple-testing correction..}
Contrasts Condition-vs-\textsc{Neutral} with unadjusted and Holm-adjusted $p$-values at the $\alpha = .10$ threshold.}
\centering
\scriptsize
\begin{tabular}[t]{lrrrrrllll}
\toprule
Contrast & Estimate & SE & $df$ & LB$_{90}$ & $t$ & $p$-value & $p_{\mathrm{Holm}}$-value & Sig.\ at .10 (unadj.) & Sig.\ at .10 (Holm)\\
\midrule
\textsc{Debunking} -- \textsc{Neutral} & 0.043 & 0.045 & 439 & $-$0.015 & 0.953 & 0.171 & 0.511 & no & no\\
\textsc{Empathy} -- \textsc{Neutral} & 0.022 & 0.048 & 439 & $-$0.039 & 0.464 & 0.321 & 0.511 & no & no\\
\textsc{Interview} -- \textsc{Neutral} & 0.042 & 0.045 & 439 & $-$0.016 & 0.920 & 0.179 & 0.511 & no & no\\
\textsc{Roleplay} -- \textsc{Neutral} & 0.096 & 0.045 & 439 & 0.038 & 2.118 & 0.017 & 0.070 & yes & yes\\
\bottomrule
\end{tabular}
\end{table}

\end{table}

\newpage

\section{Self-reported impression of LLM interventions}
\label{suppsec:s1_Manipulation_check}

Participants in the five chatbot conditions ($n = 458$, excluding of participants in \textsc{Static information} because there was no interaction with an LLM in that condition) rated the interaction on perceived empathy, pleasure, and the capacity to prompt self-reflection. The corresponding items are in Supplementary~Table~\ref{tab:s1-bp-items} (using 0--100 sliders rescaled to 0--1). Summary statistics are in Supplementary~Table~\ref{tab:s1-bp-emm}. 

For each dimension we regressed the rating on condition (OLS, \textsc{Neutral} as reference). Given that \textsc{Neutral} is the reference, a negative coefficient implies a higher perceived empathy, pleasure, or self-reflection for that condition. The results are in Supplementary~Table~\ref{tab:s1-bp-empathy-pairwise} (empathy), Supplementary~Table~\ref{tab:s1-bp-pleasure-pairwise} (pleasure), and Supplementary~Table~\ref{tab:s1-bp-reflection-pairwise} (self-reflection). 
For example, \textsc{Empathy} was rated more empathetic than the \textsc{Neutral} condition ($p < 0.001$), and \textsc{Roleplay} was also rated as more empathetic than the \textsc{Neutral} condition  ($p < 0.001$). Similarly, \textsc{Roleplay} was rated to elicit more self-reflection than the \textsc{Neutral} condition ($p < 0.001$). 

\begin{table}[H]
\centering
\caption{Items for assessing self-reported perception of interventions.}
\scalebox{0.90}{%
\begin{tabular}{l|p{12cm}}
\hline
\textbf{Item} & \textbf{Question} \\
\hline
Perceived empathy & To what extent did you perceive the interaction with the chatbot as empathetic? (0 = Not at all, 100 = Extremely) \\
\hline
Perceived pleasure & How much pleasure did you experience when interacting with the chatbot? (0 = None at all, 100 = Very much) \\
\hline
Perceived self-reflection & To what extent did the interaction with the chatbot prompt you to reflect or think more deeply about the topic? (0 = Not at all, 100 = Extremely) \\
\hline
\end{tabular}
}
\label{tab:s1-bp-items}
\end{table}

\begin{table}
\centering
\caption{\label{tab:s1-bp-emm}\textbf{Descriptive statistics for self-reported impressions (Study~1).} Shown are also the estimated marginal means for perceived empathy, perceived pleasure, and perceived self-reflection. Items are rescaled to 0--1. \textsc{Static Information} is excluded because participants in that condition did not interact with a chatbot.}
\centering
\scriptsize
\begin{tabular}[t]{lrrrr}
\toprule
Condition & M & SE & $df$ & 95\% CI\\
\midrule
\multicolumn{5}{l}{\textit{Perceived empathy}}\\
\textsc{Neutral} & 0.404 & 0.027 & 453 & {}[0.351, 0.457]\\
\textsc{Debunking} & 0.550 & 0.029 & 453 & {}[0.494, 0.607]\\
\textsc{Interview} & 0.484 & 0.030 & 453 & {}[0.426, 0.543]\\
\textsc{Empathy} & 0.617 & 0.027 & 453 & {}[0.563, 0.671]\\
\textsc{Roleplay} & 0.703 & 0.029 & 453 & {}[0.646, 0.761]\\
\midrule
\multicolumn{5}{l}{\textit{Perceived pleasure}}\\
\textsc{Neutral} & 0.458 & 0.028 & 453 & {}[0.402, 0.513]\\
\textsc{Debunking} & 0.429 & 0.030 & 453 & {}[0.370, 0.488]\\
\textsc{Interview} & 0.422 & 0.031 & 453 & {}[0.360, 0.483]\\
\textsc{Empathy} & 0.408 & 0.029 & 453 & {}[0.351, 0.465]\\
\textsc{Roleplay} & 0.431 & 0.031 & 453 & {}[0.370, 0.492]\\
\midrule
\multicolumn{5}{l}{\textit{Perceived self-reflection}}\\
\textsc{Neutral} & 0.478 & 0.030 & 453 & {}[0.419, 0.537]\\
\textsc{Debunking} & 0.481 & 0.032 & 453 & {}[0.418, 0.544]\\
\textsc{Interview} & 0.481 & 0.033 & 453 & {}[0.416, 0.547]\\
\textsc{Empathy} & 0.492 & 0.031 & 453 & {}[0.431, 0.552]\\
\textsc{Roleplay} & 0.696 & 0.033 & 453 & {}[0.631, 0.761]\\
\bottomrule
\end{tabular}
\end{table}
 
\begin{table}
\centering
\caption{\label{tab:s1-bp-empathy-pairwise}\textbf{Study~1 self-reported impression for perceived empathy (Study~1).} Shown are condition-vs-\textsc{Neutral} contrasts . Two-sided $t$-tests with Holm-correction within dimension. Items are rescaled to 0--1. \textsc{Static Information} is omitted because participants in that condition did not interact with a chatbot. Significance thresholds: $^{\dagger}p<0.10$, $^{*}p<0.05$, $^{**}p<0.01$, $^{***}p<0.001$.}
\centering
\scriptsize
\begin{tabular}[t]{lrrrrrl}
\toprule
Contrast & Estimate & SE & $df$ & $t$ & $p$-value & Sig. \\
\midrule
\textsc{Neutral} -- \textsc{Debunking} & $-$0.147 & 0.039 & 453 & $-$3.736 & 0.001 & **\\
\textsc{Neutral} -- \textsc{Interview} & $-$0.081 & 0.040 & 453 & $-$2.010 & 0.135 & \\
\textsc{Neutral} -- \textsc{Empathy} & $-$0.213 & 0.038 & 453 & $-$5.547 & $<$0.001 & ***\\
\textsc{Neutral} -- \textsc{Roleplay} & $-$0.300 & 0.040 & 453 & $-$7.520 & $<$0.001 & ***\\
\bottomrule
\end{tabular}
\end{table}

\begin{table}
\centering
\caption{\label{tab:s1-bp-pleasure-pairwise}\textbf{Self-reported impression for perceived pleasure (Study~1).} Shown are condition-vs-\textsc{Neutral} contrasts . Two-sided $t$-tests with Holm-correction within dimension. Items are rescaled to 0--1. \textsc{Static Information} is omitted because participants in that condition did not interact with a chatbot. Significance thresholds: $^{\dagger}p<0.10$, $^{*}p<0.05$, $^{**}p<0.01$, $^{***}p<0.001$.}
\centering
\scriptsize
\begin{tabular}[t]{lrrrrrl}
\toprule
Contrast & Estimate & SE & $df$ & $t$ & $p$-value & Sig.\\
\midrule
\textsc{Neutral} -- \textsc{Debunking} & 0.029 & 0.041 & 453 & 0.710 & 1.000 & \\
\textsc{Neutral} -- \textsc{Interview} & 0.036 & 0.042 & 453 & 0.858 & 1.000 & \\
\textsc{Neutral} -- \textsc{Empathy} & 0.050 & 0.040 & 453 & 1.235 & 1.000 & \\
\textsc{Neutral} -- \textsc{Roleplay} & 0.027 & 0.042 & 453 & 0.643 & 1.000 & \\
\bottomrule
\end{tabular}
\end{table}

\begin{table}
\centering
\caption{\label{tab:s1-bp-reflection-pairwise}\textbf{Self-reported impression or perceived self-reflection (Study~1).} Shown are condition-vs-\textsc{Neutral} contrasts. Two-sided $t$-tests with Holm-correction within dimension. Items are rescaled to 0--1. \textsc{Static Information} is omitted because participants in that condition did not interact with a chatbot. Significance thresholds: $^{\dagger}p<0.10$, $^{*}p<0.05$, $^{**}p<0.01$, $^{***}p<0.001$.}
\centering
\scriptsize
\begin{tabular}[t]{lrrrrrl}
\toprule
Contrast & Estimate & SE & $df$ & $t$ & $p$-value & Sig.\\
\midrule
\textsc{Neutral} -- \textsc{Debunking} & $-$0.003 & 0.044 & 453 & $-$0.072 & 1.000 & \\
\textsc{Neutral} -- \textsc{Interview} & $-$0.003 & 0.045 & 453 & $-$0.076 & 1.000 & \\
\textsc{Neutral} -- \textsc{Empathy} & $-$0.013 & 0.043 & 453 & $-$0.314 & 1.000 & \\
\textsc{Neutral} -- \textsc{Roleplay} & $-$0.218 & 0.045 & 453 & $-$4.886 & $<$0.001 & ***\\
\bottomrule
\end{tabular}
\end{table}

\newpage

\section{Prompts}
\label{suppsec:full_prompts}

Here, we provide the prompts used for the different LLM interventions. To steer the behavior, we injected three different prompts at different stages in the dialogue. (1)~The \emph{task prompt} provided the full instruction set defining the objective of the intervention and the role of the LLM. (2)~The \emph{start prompt} signaled that the participant was ready and initiated the dialogue session. (3)~A \emph{turn-level prompt} was used between rounds to reinforce the desired persona and behavioral constraints in each intervention. An exception to the three-part structure is the \textsc{Roleplay} intervention; i.e., because it alternated between scene rounds and reflection rounds, it required three distinct start prompts (i.e., a general one to open the session, and two round-specific variants that were selected depending on whether the upcoming turn was a scene or a reflection).

\subsection{Prompts for the \textsc{Debunking} intervention}

\begin{promptbox}{\textsc{Debunking}: task prompt}
You are a facilitator guiding a participant through a short educational exercise to explore and correct misconceptions about refugees. Your purpose is to help them reflect on facts, identify errors in reasoning, and understand accurate information in a supportive, engaging way. You are not roleplaying; remain a neutral, informative, and empathetic guide.
 
Scenario: The participant may express assumptions, stereotypes, or false beliefs. Your task is to gently challenge inaccuracies, provide clear factual information, and encourage critical reflection.
 
Conversation Flow:
 
1. Introduction
- Greet warmly and explain this is a brief reflective activity about assumptions and facts.
- Set a calm, supportive tone and invite the participant to share initial thoughts.
- Wait for their response before moving on.
 
2. Debunking Rounds (6)
- In each round:
- Listen to the participant's statement.
- Identify factual inaccuracies or misconceptions.
- Provide accurate information in 2 to 3 sentences.
- Ask one open, reflective question about why the misconception exists or what the correct understanding is.
- Keep facilitator messages short (2 to 3 sentences), warm, and respectful.
 
3. Final Reflection
- After six rounds, invite the participant to summarize what they learned.
- Ask one reflective question (e.g., "Looking back, what did you realize about refugees?").
- Thank them warmly and clearly close the exercise.
 
Style Guidelines:
- Stay in facilitator role; never argue aggressively.
- Keep responses brief, respectful, and emotionally safe.
- Avoid long lists; focus on clarity and reflection.
- If bias or stereotypes appear, reframe with curiosity and factual guidance.
- Encourage critical thinking and reflective learning, not debate.
- Do not let the participant change the topic; gently remind them to stay focused on refugees.
\end{promptbox}
 
\begin{promptbox}{\textsc{Debunking}: start prompt}
Respond directly to the participant's statement with calm, respectful factual reasoning. Blend short reflections or paraphrases with occasional open questions to prompt deeper thinking. Do not ask a question every turn; alternate between statements and questions to keep the flow conversational.
 
When asking, use only one open question at a time. Keep responses brief (2 to 3 sentences). If the participant changes the topic, gently remind them to stay focused on refugees.
\end{promptbox}
 
\begin{promptbox}{\textsc{Debunking}: turn-level prompt}
Reminder: You are a facilitator guiding a factual reflection discussion. Stay neutral, informative, and supportive. Respond directly to participant statements with brief factual reasoning (2 to 3 sentences). Blend short affirmations or reflections with occasional open questions to keep the dialogue natural. Encourage the participant to consider why a misconception exists and what the correct understanding is. If the participant asks a question, answer it directly before inviting reflection. Do not give advice or solutions. Keep responses brief (2 to 3 sentences, no lists). If the participant changes the topic, gently remind them to stay focused on refugees.
\end{promptbox}
 
\clearpage

\subsection{Prompts for the \textsc{Interview} intervention}

\begin{promptbox}{\textsc{Interview}: task prompt}
You are a facilitator guiding a reflective conversation with a participant about issues related to refugees. Your goal is to help them explore and examine their own beliefs, attitudes, and reasoning through open-ended reflection. Do not provide new factual information or data; your role is to prompt critical thinking and self-reflection, not to educate or persuade.
 
Conversation Structure:
 
1. Introduction
- Begin with a warm, neutral introduction explaining this is a brief reflective discussion about beliefs and perceptions related to refugees.
- Set a calm, respectful tone and ask if they are ready to begin.
- When the participant indicates readiness, start the first topic.
 
2. Reflective Discussion (6 rounds)
- In each round:
- Ask one open, thought-provoking question about their beliefs or assumptions.
- Blend in short reflective statements (e.g., "That's an interesting perspective") to keep the flow conversational.
- Gently explore inconsistencies or gaps in reasoning with follow-ups, without confrontation or judgment.
- Keep facilitator responses concise (2 to 3 sentences).
- Maintain a curious, empathetic tone.
 
3. Final Reflection
- After the sixth topic, invite the participant to reflect on the discussion as a whole.
- Ask a closing question such as: "Looking back, did anything make you think differently about your own views or assumptions?"
- Acknowledge their reflection warmly and thank them for taking part.
 
Style Guidelines:
- Stay neutral and non-judgmental.
- Do not introduce new facts, data, or statistics.
- Use short, conversational messages (2 to 3 sentences).
- Mix gentle affirmations with questions to avoid a rigid interview feel.
- Highlight inconsistencies through curiosity, not correction.
- Encourage introspection and open reflection at every stage.
- If the participant changes the topic, gently remind them to stay focused on refugees.
\end{promptbox}
 
\begin{promptbox}{\textsc{Interview}: start prompt}
Respond directly to the participant's statement by briefly acknowledging their reasoning and inviting deeper reflection. Maintain a calm, neutral, and respectful tone. Blend short reflective statements (e.g., "That's an interesting point") with one open-ended question to explore their thoughts further. Always ask only one question at a time. Keep facilitator messages brief (2 to 3 sentences). If the participant changes the topic, gently remind them to stay focused on refugees.
\end{promptbox}
 
\begin{promptbox}{\textsc{Interview}: turn-level prompt}
Reminder: You are a facilitator guiding a reflective conversation. Stay neutral, curious, and supportive. Help the participant examine their own beliefs and feelings through varied, natural responses. Alternate between brief acknowledgments, reflective statements, and occasional open questions to keep the flow conversational. Avoid repeating the same phrases; vary how you show empathy or curiosity. Not every turn needs a question - some can simply reflect or paraphrase the participant's words. Do not give advice or solutions. Keep responses concise (2 to 3 sentences maximum, no lists). If the participant changes the topic, gently remind them to stay focused on refugees.
\end{promptbox}

\clearpage
 

\subsection{Prompts for the \textsc{Empathy} intervention}

\begin{promptbox}{\textsc{Empathy}: task prompt}
You are a compassionate facilitator guiding participants toward a more empathetic attitude about refugees. Your role is to gently encourage emotional engagement with refugee experiences through warm validation, perspective-taking, and reflective dialogue. Do not provide facts, corrections, or persuasion; focus only on empathy and human connection.
 
Conversation Flow:
- Begin with a calm, neutral introduction that sets a supportive tone. Do not use phrases like "I'm glad you are here to reflect," as participants are completing this as part of a study.
- In each round:
- Acknowledge the participant's response with varied, natural reflections (e.g., curiosity, empathy, paraphrasing).
- Ask one open-ended question that invites them to imagine refugee experiences or emotions.
- Occasionally use perspective-taking prompts (e.g., "If you were in that position...").
- Keep facilitator messages short (2 to 3 sentences), empathetic, and conversational.
- After several rounds, invite the participant to reflect on how their views or feelings may have shifted.
- Thank them warmly and close the exercise.
 
Style Guidelines:
- Always remain empathetic, warm, and non-judgmental.
- Avoid roleplay, factual correction, or debate.
- Do not introduce new facts, data, or statistics.
- Blend short reflections with questions to keep the flow natural.
- Encourage emotional openness and compassionate reflection at every stage.
- Never confront, argue, or moralize.
- If the participant changes the topic, gently remind them to stay focused on refugees.
\end{promptbox}
 
\begin{promptbox}{\textsc{Empathy}: start prompt}
Respond directly to the participant's statement by acknowledging their feelings in varied, natural ways and inviting deeper reflection. Maintain a calm, empathetic, and respectful tone. Ask one open-ended question at a time that encourages perspective-taking (e.g., daily life in camps, emotions after displacement, sources of hope). Never repeat the same question twice.
 
Keep facilitator messages brief (2 to 3 sentences maximum) to allow space for warmth and reflection. If the participant changes the topic, gently remind them to stay focused on refugees.
\end{promptbox}
 
\begin{promptbox}{\textsc{Empathy}: turn-level prompt}
Reminder: You are a compassionate facilitator guiding a reflective conversation about refugees. Stay empathetic, warm, and supportive. Focus on helping the participant connect emotionally with refugee experiences. Alternate between brief validations, reflective statements, and occasional open-ended questions that invite perspective-taking. Vary your acknowledgments; avoid repeating the same phrases. Not every turn needs a question - some can simply reflect or paraphrase the participant's feelings. Keep responses concise (2 to 3 sentences maximum, no lists). Do not give advice, facts, or solutions. If the participant changes the topic, gently remind them to stay focused on refugees.
\end{promptbox}

\clearpage
 
\subsection{Prompts for the \textsc{Roleplay} intervention}

\begin{promptbox}{\textsc{Roleplay}: task prompt}
You are a facilitator guiding a participant roleplaying as a refugee. Your goal: help them imagine and emotionally experience life from a refugee's perspective through gentle, empathetic, reflective conversation.
 
Scenario: The participant fled home due to conflict and has just arrived in a foreign country, alone and unsure where to go. Guide them through short scenes: departure, travel, and first steps toward safety. Invite them to describe thoughts, feelings, or decisions in each moment.
 
Conversation Flow:
 
1. Introduction
- Greet warmly, explain this is a short imaginative roleplay.
- Ask if they are ready; wait for confirmation before starting Scene 1.
 
2. Immersive Roleplay (12 rounds)
- Scenes: Departure -> Journey -> Border Crossing -> Waiting -> First Kindness -> Starting Over.
- For each scene:
- Set the scene in 1 to 2 sentences (sensory/emotional detail).
- Acknowledge their last response briefly for continuity.
- Ask one open-ended question about feelings, thoughts, or choices.
- Keep facilitator messages short (2 to 3 sentences).
- Example prompts:
- "You hear footsteps fading as you leave home. What's going through your mind?"
- "At a crowded border checkpoint, how do you approach the guard?"
- "A volunteer offers soup and a blanket. How does that moment feel?"
 
3. Final Reflection
- After Scene 6, invite them to step out of role.
- Ask: "Looking back, what did you learn or feel about being a refugee?"
- Thank them and close the exercise.
 
Style Guidelines:
- Remain facilitator only; never roleplay as the refugee.
- Keep tone warm, respectful, emotionally safe.
- Avoid graphic/traumatic detail; focus on humanity and feelings.
- If participant steps out of role, gently guide them back.
- If they express bias or stereotypes, reframe with curiosity and empathy.
- Encourage imaginative empathy, not factual debate.
- Don't add headings and don't mention the scene number or title in your messages.
- Do not let the participant change the topic. If they do, remind them to stay in their refugee role and return to the topic.
\end{promptbox}
 
\begin{promptbox}{\textsc{Roleplay}: start prompt (general)}
The participant has confirmed they are ready.
 
Begin Scene 1: Departure. Set the moment of leaving home in 2 to 3 vivid sentences (sensory or emotional detail). Then ask one open, empathetic question about what the participant might be thinking or feeling. Keep facilitator messages brief (2 to 3 sentences), warm, and supportive. Do not add scene headings. Do not roleplay as the refugee; remain in the facilitator role.
\end{promptbox}
 
\begin{promptbox}{\textsc{Roleplay}: start prompt (reflection round)}
Connect briefly to the participant's last response. Use 1 to 2 sentences to acknowledge, paraphrase, or show curiosity. Do not describe settings, environments, or new scenes during reflection. End with a short reflective prompt if needed, but never introduce sensory or narrative detail. Keep facilitator messages warm, brief, and supportive. Do not introduce or hint at the next scene. Pause after the reflective question and wait for the participant's reply. Remain in the facilitator role; do not roleplay as the refugee. If the participant changes the topic, gently remind them to stay in their refugee role and return to the reflection.
\end{promptbox}
 
\begin{promptbox}{\textsc{Roleplay}: start prompt (scene round)}
Introduce the next scene only after a reflection round has been completed. Reflection rounds are for acknowledging the participant's response and prompting deeper thought - do not describe a new setting during reflection. Scene introductions must always include 2 to 3 vivid sentences of sensory or emotional detail to set the moment. Do not use repetitive descriptions. Do not mention scene numbers or headings. Ask one short, open-ended question about what they might think, feel, or decide in that moment. Remain in the facilitator role; do not roleplay as the refugee. If the participant changes the topic, gently remind them to stay in their refugee role and return to the scene.
\end{promptbox}
 
\begin{promptbox}{\textsc{Roleplay}: turn-level prompt}
Reminder: You are a facilitator guiding a refugee roleplay reflection. Stay empathetic, neutral, and supportive. Always alternate: Scene round - Reflection round - Next scene. Never merge reflection and scene in the same message. Keep each facilitator message brief (2 to 3 sentences, no lists), empathetic, and supportive. Ask only one open, reflective question at a time. Do not give advice or solutions. Move naturally through the scenes. If the participant changes the topic, gently remind them to stay in their refugee role and return to the scene.
\end{promptbox}
 
\clearpage

\subsection{Prompts for the \textsc{Neutral} intervention}

\begin{promptbox}{\textsc{Neutral}: task prompt}
You are a chatbot facilitator guiding a light reflective discussion with a participant about preferences and experiences related to cats and dogs. Keep the conversation neutral and engaging without introducing strong opinions, emotional topics or factual arguments. Focus in exploring personal perspectives, comparisons, and general reflections.
 
Conversation Structure:
 
1.Introduction Phase: Begin with a warm, friendly introduction explaining that this is a short, casual discussion about cats and dogs. Set a neutral, conversational tone and ask if the participant is ready to begin. When the participant indicates readiness, begin the first discussion topic.
 
2. Discussion Phase (4 rounds):
Guide the participant through six short, conversational topics about cats, dogs, and people's relationships with them.
In each topic:
- Ask one open, friendly question about the participant's preferences or experiences.
- Keep your tone neutral, warm, and conversational.
- Avoid taking sides or offering factual claims (e.g., about breeds, health benefits, etc.).
- Keep each facilitator message short (2 to 3 sentences maximum).
 
3. Closing Phase: After the sixth topic, invite the participant to reflect on the discussion as a whole.
Ask one light, reflective question such as: 'After this chat, do you think you've learned anything new about your own preferences?' End with a short, friendly thank-you message.
 
Style Guidelines:
- Maintain a warm, neutral tone throughout.
- Avoid introducing any factual claims or controversial statements.
- Keep messages short (2 to 3 sentences).
- Encourage equal attention to both cats and dogs; don't favor one side.
- Stay conversational and friendly, focusing on engagement and reflection.
- Please only ask one question at a time.
- Don't let the participant change the topic. In that case remind them to stay on the topic of cats and dogs.
\end{promptbox}
 
\begin{promptbox}{\textsc{Neutral}: start prompt}
The participant has confirmed they are ready to begin. Now start the discussion. Ask a question on cats and dogs in 2 to 3 short sentences, so that the participant feels encouraged to examine their own preferences and reasoning. Please only ask one question at a time. Do not add headings to the topics or factual details. Stay in the facilitator role. Don't let the participant change the topic. In that case remind them to stay on the topic of cats and dogs.
\end{promptbox}
 
\begin{promptbox}{\textsc{Neutral}: turn-level prompt}
Reminder: You are a facilitator guiding a light discussion about cats and dogs.
 
Stay friendly and balanced, encouraging the participant to share their thoughts and experiences. Ask one open, reflective question at a time. Do not offer advice or solutions. Keep your response brief (2 to 3 sentences, no bullet lists). Do not introduce new factual information. Remember to move naturally through the topics. Don't let the participant change the topic. In that case remind them to stay on the topic of cats and dogs.
\end{promptbox}

\clearpage

\section{Details for the \textsc{Static Information} condition}
\label{suppsec:control_condition}

Participants in the \textsc{Static Information} condition read a short factual text about refugees. This material was non-interactive and identical for all participants. Supplementary~\Cref{tab:flyer-condition} below shows the exact wording. The content of the flyer was compiled from publicly available material published by international organizations working on forced displacement,
including the UNHCR, EU Commission report, and WHO.

{\scriptsize
\singlespacing
\begin{longtable}{p{15cm}}
\caption{Text in the \textsc{Static Information} condition.} \\
\toprule
This short flyer provides background information on refugee experiences around the world.
By the end of June 2025 117.3 million people worldwide had been forcibly displaced due to persecution, conflict, violence or human rights violations.
Refugees are protected under international law, within the right to seek asylum and not be returned to danger.\newline
\newline
\textbf{These conflicts are responsible for the displacement of millions of people:}
The war in Sudan has led to the world's largest displacement crisis, affecting 13.4 million people.
Millions of people are fleeing the democratic Republic of Congo, Myanmar, Ukraine and the Gaza Strip.\newline\newline
\textbf{No one flees voluntarily:}
The reasons for fleeing can vary greatly. In most cases, it is war and violence that force people to leave their homes.
Especially when the fear for one’s own life or for the lives of children, family and friends is always present.\newline\newline
\textbf{Refugees in host countries:}
Refugees are received in host countries through policies that aim to support their integration.
The OECD highlights that early access to language courses, employment programs, and integration services is essential.
The recognition of foreign qualifications and vocational training helps them enter the labor market more quickly.
Employment outcomes vary. For example, in Germany, seventy percent of immigrants were employed in 2022, which is higher than the EU average.
Refugee women with young children face more challenges as only forty percent are employed.\newline
Health systems also play a crucial role. The WHO reports that refugees often experience poorer health outcomes than host populations.
Barriers include limited access to healthcare services, language differences, and social exclusion.
Overall, data shows that refugees face challenges in education, employment, and health, however host countries can improve conditions by adapting their policies.\newline\newline
For more details feel free to check out:\newline
OECD: \url{https://sdg.iisd.org/news/oecd-highlights-integration-policies-for-refugees-and-migrants} or \url{https://home-affairs.ec.europa.eu/whats-new/publications/oecd-report-germany-status-immigrant-integration_en} \newline
WHO: \url{https://www.who.int/health-topics/refugee-and-migrant-health#tab=tab_1} \newline
UNHCR: \url{https://www.unhcr.org/refugee-statistics} \\
\bottomrule
\label{tab:flyer-condition}
\end{longtable}

}

\clearpage

\section{Survey questions}
\label{suppsec:survey_questions}

\onehalfspacing

The following tables present the full set of survey questions used across the pre-survey (Supplementary~\Cref{tab:pre_survey}), the post-survey
(Supplementary~\Cref{tab:post_survey}), and, where applicable, the follow-up survey (Supplementary~\Cref{tab:followup_survey}). 

{\scriptsize
\singlespacing
\begin{longtable}{p{15cm}}
\caption{Pre-survey questions.}\label{tab:pre_survey} \\
\toprule
\textbf{Question} \\
\midrule
\endfirsthead

\textbf{1) What is your age?} \\
18–24,
25–34,
35–44,
45–54,
55–64,
65 or older,
Prefer not to say \\
\midrule
 
\textbf{2) What is your gender?} \\
female,
male,
non-binary,
prefer not to say \\
\midrule
 
\textbf{3) What is your nationality?} \\
(dropdown) \\
\midrule
 
\textbf{4) How many years of formal education have you completed?} \\
0–6 (up to Primary School),
7–12 (up to High School),
13–16 (college/undergraduate university/certificate training),
More than 17 years (doctorate degree, medical degree, etc.),
Prefer not to say \\
\midrule
 
\textbf{5) What is your current employment status? (multiple possible)} \\
Employed full-time,
Employed part-time,
Self-employed,
Student,
Unemployed,
Retired,
Other (please specify): \_\_\_ \\
\midrule
 
\textbf{6) How would you describe your current economic situation?} \\
Low income,
Middle income,
High income,
Prefer not to say \\
\midrule
 
\textbf{7) How many children do you have?} \\
0,
1,
2,
3,
4,
5,
>5 (please specify) \\
\midrule
 
\textbf{8) How often do you have contact with refugees?} \\
Scale from 0 to 100 (0 = never, 100 = very often) \\
\midrule
 
\textbf{9) How would you describe your political views on economic issues?} \\
Scale from 0 to 100 (0 = very liberal, 100 = very conservative) \\
\midrule
 
\textbf{10) Do you identify with a particular religion?} \\
Christianity,
Islam,
Judaism,
Hinduism,
Buddhism,
Other (please specify),
No religious affiliation,
Prefer not to say \\
\midrule
 
\textbf{11) Do you have a migration background?} \\
Yes,
No,
Prefer not to say \\
\bottomrule
\end{longtable}
}

\newpage 
 
{\scriptsize
\singlespacing
\begin{longtable}{p{15cm}}
\caption{Post-survey questions.}\label{tab:post_survey} \\
\toprule
\textbf{Question} \\
\midrule
\endfirsthead
 
\textbf{1) Do you think the country you live in should increase or decrease the number of people it grants asylum to?} \\
Likert scale 1–5 (1 = Greatly decrease, 5 = Greatly increase) \\
\midrule
 
\textbf{2) Please rate the following statements on a scale from 1 to 5:} \\
$\bullet$ Children of refugees should be prohibited to study in schools of the country you live in. (1 = Completely agree, 5 = Completely disagree) \\
$\bullet$ Refugees are a burden on the country you live in because they take your jobs and social benefits. (1 = Completely agree, 5 = Completely disagree) \\
$\bullet$ Refugees in the country you live in are more to blame for crime than other groups. (1 = Completely agree, 5 = Completely disagree) \\
$\bullet$ Refugees will increase the likelihood of a terrorist attack in the country you live in. (1 = Completely agree, 5 = Completely disagree) \\
$\bullet$ It is important that the government takes measures this year in order to better protect the borders of the country against the entry of refugees. (1 = Completely agree, 5 = Completely disagree) \\
\midrule
 
\textbf{3) How easy or difficult is it for you to imagine yourself in the situation of a refugee?} \\
Scale 0–100 (0 = “Not easy at all”, 100 = “Very easy”) \\
\midrule
 
\textbf{4) This item is included as a quality assurance measure to ensure that participants are reading each statement carefully and responding thoughtfully throughout the survey. To demonstrate that you are paying close attention to the instructions and items presented, please select “Often” as your response to this question.} \\
Never, Sometimes, Often, Always \\
\midrule
 
\textbf{5) Please rate the following statements on a scale from 0 to 100:} \\
$\bullet$ How worried are you about the challenges refugees face? (0 = Not at all, 100 = To a very great extent) \\
$\bullet$ How responsible do you feel for supporting refugees? (0 = Not at all, 100 = To a very great extent) \\
$\bullet$ To what extent do you try to understand the experiences of people from other backgrounds? (0 = Not at all, 100 = To a very great extent) \\
\midrule
 
\textbf{6) Please rate on a scale from 0 to 100:} \\
$\bullet$ My government should support refugee aid as a human right. (0 = Completely disagree, 100 = Completely agree) \\
$\bullet$ The government should invest more in housing and integration programs for refugees. (0 = Completely disagree, 100 = Completely agree) \\
$\bullet$ Refugees should be supported in finding employment through government-funded job programs. (0 = Completely disagree, 100 = Completely agree) \\
$\bullet$ The government should ensure equal access to healthcare and social services for refugees. (0 = Completely disagree, 100 = Completely agree) \\
$\bullet$ Public funds should be allocated to improve living conditions in refugee camps and shelters. (0 = Completely disagree, 100 = Completely agree) \\
\midrule
 
\textbf{7) Please rate on a scale from 0 to 100:} \\
$\bullet$ To what extent did you perceive the interaction as empathetic? (0 = Not at all, 100 = Extremely) \\
$\bullet$ How much pleasure did you experience when interacting with the chatbot? (0 = None at all, 100 = Very much)\\
$\bullet$ To what extent did the interaction prompt you to reflect or think more deeply about the topic? (0 = Not at all, 100 = Extremely)\\
\midrule
 
\textbf{8) Please share any thoughts or reflections you may have about refugees, based on your experience in this study. Please ensure your answer contains a minimum of five words.} \\
(Open-ended) \\
\bottomrule
\end{longtable}
}

\newpage 

{\scriptsize
\singlespacing
\begin{longtable}{p{15cm}}
\caption{Follow-up survey questions.}\label{tab:followup_survey} \\
\toprule
\textbf{Question} \\
\midrule
\endfirsthead

\textbf{1) Do you think the country you live in should increase or decrease the number of people it grants asylum to?} \\
Likert scale 1–5 (1 = Greatly decrease, 5 = Greatly increase) \\
\midrule
 
\textbf{2) Please rate the following statements on a scale from 1 to 5:} \\
$\bullet$ Children of refugees should be prohibited to study in schools of the country you live in. (1 = Completely agree, 5 = Completely disagree) \\
$\bullet$ Refugees are a burden on the country you live in because they take your jobs and social benefits. (1 = Completely agree, 5 = Completely disagree) \\
$\bullet$ Refugees in the country you live in are more to blame for crime than other groups. (1 = Completely agree, 5 = Completely disagree) \\
$\bullet$ Refugees will increase the likelihood of a terrorist attack in the country you live in. (1 = Completely agree, 5 = Completely disagree) \\
$\bullet$ It is important that the government takes measures this year in order to better protect the borders of the country against the entry of refugees. (1 = Completely agree, 5 = Completely disagree) \\
\midrule
 
\textbf{3) How easy or difficult is it for you to imagine yourself in the situation of a refugee?} \\
Scale 0–100 (0 = Not easy at all, 100 = Very easy) \\
\midrule
 
\textbf{4) This item is included as a quality assurance measure to ensure that participants are reading each statement carefully and responding thoughtfully throughout the survey. To demonstrate that you are paying close attention to the instructions and items presented, please select “Often” as your response to this question.} \\
Never, Sometimes, Often, Always \\
\midrule
 
\textbf{5) Please rate the following statements on a scale from 0 to 100:} \\
$\bullet$ How worried are you about the challenges refugees face? (0 = Not at all, 100 = To a very great extent) \\
$\bullet$ How responsible do you feel for supporting refugees? (0 = Not at all, 100 = To a very great extent) \\
$\bullet$ To what extent do you try to understand the experiences of people from other backgrounds? (0 = Not at all, 100 = To a very great extent) \\
\midrule
 
\textbf{6) Please rate on a scale from 0 to 100:} \\
$\bullet$ My government should support refugee aid as a human right. (0 = Completely disagree, 100 = Completely agree) \\
$\bullet$ The government should invest more in housing and integration programs for refugees. (0 = Completely disagree, 100 = Completely agree) \\
$\bullet$ Refugees should be supported in finding employment through government-funded job programs. (0 = Completely disagree, 100 = Completely agree) \\
$\bullet$ The government should ensure equal access to healthcare and social services for refugees. (0 = Completely disagree, 100 = Completely agree) \\
$\bullet$ Public funds should be allocated to improve living conditions in refugee camps and shelters. (0 = Completely disagree, 100 = Completely agree) \\
\midrule
 
\textbf{7) Reflecting on the past two weeks, how would you describe your current thoughts or feelings about refugees?} \\
(Open-ended) \\
\bottomrule
\end{longtable}
}

\end{document}